\documentclass[10pt]{article}
\usepackage[margin=1in]{geometry}
\usepackage{mathpazo}
\usepackage{booktabs}
\usepackage{afterpage}
\usepackage{float}

\usepackage{bm}
\usepackage{amsmath,amssymb,amsthm,amsfonts}
\usepackage{xspace}
\usepackage{graphicx}
\usepackage{color}
\usepackage[colorlinks=true,linkcolor=blue,citecolor=blue,urlcolor=blue]{hyperref}
\usepackage{todonotes}
\usepackage{subcaption}
\usepackage{comment}
\usepackage[ruled]{algorithm2e}
\usepackage[T1]{fontenc}
\usepackage[backend=biber,style=alphabetic,natbib=true,maxbibnames=99,minalphanames=3,maxcitenames=2]{biblatex}
\usepackage{makecell}

\usepackage{array}
\usepackage{textcomp}
\usepackage{stfloats}
\usepackage{url}
\usepackage{verbatim}

\usepackage{listings}

\usepackage{xcolor}
\usepackage{courier}
\usepackage{overpic}

\definecolor{codegreen}{rgb}{0,0.6,0}
\definecolor{codegray}{rgb}{0.5,0.5,0.5}
\definecolor{codepurple}{rgb}{0.58,0,0.82}
\definecolor{backcolour}{rgb}{0.95,0.95,0.92}

\lstdefinestyle{mystyle}{
    backgroundcolor=\color{backcolour},
    commentstyle=\color{codegreen},
    keywordstyle=\color{magenta},
    stringstyle=\color{codepurple},
    basicstyle=\ttfamily\footnotesize,
    breakatwhitespace=false,
    breaklines=true,
    captionpos=b,
    keepspaces=false,
    showspaces=false,
    showstringspaces=false,
    showtabs=false,
    tabsize=2
}

\title{SPAIC: A Spike-based Artificial Intelligence Computing Framework}
\author{Chaofei Hong \\
	   Zhejiang Lab\\
 \and
	   Mengwen Yuan \\
	   Zhejiang Lab 
\and 
	   Mengxiao Zhang \\ 
	   Zhejiang Lab 
\and 
	   Xiao Wang \\
	   Zhejiang Lab 
\and 
	   Chengjun Zhang \\
	   Zhejiang Lab 
\and 
        Jiaxin Wang \\
	   Zhejiang Lab \\
\and 
        Gang Pan\\
        Zhejiang University \\
 \and
        Zhaohui Wu\\
	   Zhejiang University \\
\and 
        Huajin Tang \thanks{Corresponding author: Huajin Tang. (e-mail:htang@zju.edu.cn)}\\
	   Zhejiang University
}
\makeatletter
\renewcommand{\paragraph}{%
  \@startsection{paragraph}{4}%
  {\z@}{1.5ex \@plus 1ex \@minus .2ex}{-1em}%
  {\normalfont\normalsize\bfseries}%
}
\makeatother

\date{}

\begin{document}
    \maketitle
\begin{abstract}

Neuromorphic computing is an emerging research field that aims to develop new intelligent systems by
integrating theories and technologies from multi-disciplines such as neuroscience and deep learning.
Currently, there have been various software frameworks developed for the related fields, but there is a lack of an efficient
framework dedicated for spike-based computing models and algorithms.
In this work, we present a Python based spiking neural network (SNN) simulation and training framework, aka \texttt{SPAIC} that aims to
support brain-inspired model and algorithm researches integrated with features from both deep learning and neuroscience.
To integrate different methodologies from the two overwhelming disciplines, and balance between flexibility and
efficiency, \texttt{SPAIC} is designed with neuroscience-style frontend and deep learning backend structure.
We provide a wide range of examples including neural circuits simulation, deep SNN learning and neuromorphic applications, demonstrating the concise coding style and wide usability of our framework.
The \texttt{SPAIC} is a dedicated spike-based artificial intelligence computing platform, which will significantly facilitate the design, prototype and validation of new models, theories and applications. Being user-friendly, flexible and high-performance, it will help accelerate the rapid growth and wide applicability of neuromorphic computing research.

\end{abstract}

\section{Introduction}

Recent advances from multiple research fields bring us closer to understanding and mimicking the brain.
In neuroscience, accumulating models and theories at different scales are developed to explore and understand the neural systems.
In another approach, the field of deep learning, which aims to replicate brain functions with simplified models, has
made tremendous progress in recent years\cite{yan2015deep}.
Even though those two fields are started with a similar goal, they have revealed quite different answers for the same
question.
In this circumstance, the idea of integrating neural mechanisms found in neuroscience with machine learning techniques
to develop brain-inspired computing systems with higher intelligent functions or better computational efficiency has
attracted growing attentions \cite{abbott2008theoretical, park2013structural, marblestone2016toward}.
Those researches are growing into a new multi-discipline filed, often called brain-inspired computing or neuromorphic
computing.

The efficient and easily accessible software tools have largely accelerated the development of
deep learning \cite{abadi2016tensorflow,paszke2019pytorch}.
Similarly, the development of computational neuroscience have also accompanied with a series of specialized simulation
tools.
However, current software tools in both deep learning and neuroscience field are not designed for this new
interdisciplinary research.

As those two fields have developed separate ways of modeling and studying neural models,
the lack of a generally compatible training and simulation framework makes it harder for researchers to integrate
theories from multiple disciplines and quickly validate new ideas.
To meet this demand, we present the \texttt{SPAIC} (Spike-based Artificial Intelligence Computing), a Python based spiking neural
network (SNN) training and simulation framework, which blend the programing style and techniques from both deep learning and neuroscience,
and provide a platform to easily test brain-inspired mechanisms and theories in learnable neuromorphic models
\cite{schuman2022opportunities}.

SNN is a common tool in computational neuroscience for modeling neural systems in a bottom-up approach.
Compared to artificial neural networks (ANNs), SNNs captured key computational properties of the biological neural
system with dynamical computation models.
Researchers have used SNNs to study mechanisms underlying various neural dynamical behaviors, explore the functional role
of different neuronal arrangements and connectivity patterns, and simulate the brain activities in high level of
realisticity \cite{fan2019brief}.
With improving understanding of the brain, it is expected that such biologically realistic model can come closer to
achieving brain's remarkable intelligence abilities than more abstract models.
Several SNN simulation tools such as \texttt{Neuron} \cite{carnevale2006neuron}, \texttt{Brian2}
\cite{stimberg2019brian}
and \texttt{Nest} \cite{eppler2009pynest} have been developed, which can help researchers construct network models with
customized neural dynamics and simulate neural activities with high precision.
However, those software tools are not optimized for building complex intelligent models, especially those based on big
data and training algorithms, and their computational efficiency is lower than current deep learning frameworks.
One key objective of our framework is to support building biologically realistic SNN models with high flexibility and
efficiency.
To achieve this goal, the \texttt{SPAIC} framework is designed with a frontend-backend architecture which decouples neural
model creation and simulation.
In the frontend, the \texttt{SPAIC} provides user-friendly interfaces which can hierarchically compose complex network
structures with neuron assemblies and various connection policies, and flexibly define neural dynamics with customizable neuron
models, synapses and learning rules.
In the backend, frontend network models are built into an optimized computation graph, and run by a simulator engine based
on popular deep learning frameworks, such as \texttt{PyTorch} \cite{paszke2019pytorch} and \texttt{TensorFlow}
\cite{abadi2016tensorflow},
which provide efficient CPU/GPU parallel computing and autograd techniques.

In another approach, cognitive neuroscience decomposes complex cognitive processes into elementary computational
components with associated brain regions and brain activities, and gives a top-down view of the brain's functional
organization \cite{gazzaniga2009cognitive}.
The top-down understanding of the brain provides an essential guidance for building brain-inspired models.
Recent trends witness the adoption of mathematical modeling in cognitive neuroscience, which bridges the gap between
experimental research and computational work \cite{kriegeskorte2018cognitive, palmeri2017model}.
Those researches often focused on brain states and information transfer in larger scales with higher-level models.
Hence, our framework also aims to support higher-level models such as ANNs and mean-field neural models, and those
different level of models can be built separately or merged into one hybrid network model.
In this way, researchers can easily build and test their ideas with more abstract models, and then those results can
be used as a guidance for detailed SNN modeling.

Given the computation components and functional organization of neural systems, a biologically reasonable brain-inspired
network model can be constructed.
However, it is difficult to manually design the high-dimensional parameters of the model to achieve complex brain functions.
Building a functional brain-inspired AI model still needs optimization technologies.
In recent years, advance in deep learning provides techniques and theories for optimizing complex network structures.
Another key objective of our framework is to naturally blend the understanding and techniques in the fields of deep
learning, cognitive neuroscience and computational neuroscience into one unified modeling procedure.
However, incorporating deep learning methodologies, such as gradient descent, into SNN models remains challenging,
both in theory and software engineering.
Gradient propagation is one of the fundamental techniques in deep learning, which is usually incompatible with the complex
neural dynamics necessary for brain functions.
Moreover, biological neural systems exhibit much more diverse learning mechanisms, such as timing-based or rate-based
local learning rules \cite{caporale2008spike}, dopamine driven learning \cite{wise2004dopamine}, and wiring/rewiring
mechanisms \cite{chklovskii2004cortical} which are not in line with the gradient descent optimization methodology.
Although numerous works have demonstrated that SNNs can be efficiently trained using surrogate gradient algorithms
\cite{neftci2019surrogate},
the inclusion of more complex structure and neural dynamics to achieve superior performance remains a major obstacle.
Moreover, biological neural systems have many distinct features such as sparse connectivity, spatial structure, nonlinear
neuronal and synaptic dynamics, spontaneous activity and information transmission delays, which could be essential for
neural computation.
The community is calling for new learning rules that combine the learning efficiency of deep learning with biological features
of neural networks.
To help researchers in this endeavor, the \texttt{SPAIC} designed a learner class that defines learning algorithms in a general
training framework, which support both gradient-based learning and local (plasticity based) learning rules.
And learner class provides interfaces to flexibly modify any given part of SNN computation, such as computation in
connection, neuron model, and synapses.
Some classical learning algorithms have been implemented in the framework, while it is encouraged that users can develop
their own algorithms through this framework.

The \texttt{SPAIC} framework is an open-source project that is under intensive development, this paper is based on the primary release
version.
The source code of the framework is available at https://github.com/ZhejianglabNCRC/SPAIC, the documentation of the framework can be found
at https://spaic.readthedocs.io, and the source code for all the examples in this article has been deposited at https://github.com/ZhejianglabNCRC/SPAIC\_paper\_examples.
The paper is structured as follows: In Section 2, we discussed existing software tools related to neural network
modeling
in the fields of deep learning, neuroscience and neuromorphic computing, and discussed why we need a new framework.
In Section 3, the framework structure and usage procedure is described in detail, where the motivation and functionality
of each software component are explained.
Example codes and case studies are given in Section 4 to demonstrate the coding style and potential usage situations of
the \texttt{SPAIC}.
Then, future developments of \texttt{SPAIC} are discussed in Section 5.

\section{Related Work}
There are a number of software tools can be used for modeling neural systems.
Each is tailored toward specific application domains.
According to the characteristics of these software tools, they can be roughly divided into three categories: deep
learning, computational neuroscience and neuromorphic computing frameworks.
The computational neuroscience frameworks focus on simulating the details of physiological structure and neurons while
deep learning frameworks focus on achieving brain functions using simplified models and learning algorithms.
Neuromorphic computing frameworks need to combine the above two types of frameworks for constructing model with brain-like
functional and physiological details.
In this section, we describe the relevant software tools and the challenges.
Some popular frameworks are compared in Table \ref{FrameworkComparison}.

\subsection{Deep Learning Frameworks}
In recent years, a number of deep learning frameworks have emerged, such
as \texttt{Caffe} \cite{jia2014caffe}, \texttt{Theano} \cite{al2016theano}, \texttt{PyTorch}
\cite{paszke2019pytorch}, \texttt{TensorFlow} \cite{abadi2016tensorflow}, etc.
They focus on the development, training and inference of deep learning networks.
Among them, \texttt{PyTorch} and \texttt{TensorFlow} are the two most popular open source libraries for machine
learning to date.
For \texttt{TensorFlow}, the first version (\texttt{TensorFlow 1.x}) only supports static computation graphs whereas
\texttt{TensorFlow 2.0} supports dynamic models and ease the process of building machine learning frameworks.
It is well suited for large-scale distributed training and can run on CPUs, GPUs, or large-scale
distributed systems.
\texttt{PyTorch} provides a Pythonic programming style and supports dynamic tensor computations with automatic
differentiation and strong GPU acceleration.
Its fast performance is achieved by being written mostly in C++.
It is easy for users to develop, debug, and run neural networks.
In a word, these frameworks implement function by optimizing complex network based on simple mathematical model.
They can not be used to develop and train SNN directly.

\subsection{Computational Neuroscience Frameworks}
According to the needs of neuroscience theoretical simulation, several simulation software tools have emerged, mainly
including \texttt{NEURON} \cite{hines2001neuron}, \texttt{GENSIS} \cite{cornelis2012Genesis}, \texttt{CARLsim}
\cite{chou2018carlsim}, \texttt{NEST} \cite{Gewaltig2007Nest}, \texttt{Brian2} \cite{stimberg2019brian}, and
\texttt{BrainPy} \cite{wang2021BrianPy}, which achieve biological realism in different levels.
\texttt{NEURON} and \texttt{GENSIS} focus on simulating detailed realistic biological neuron with properties that
include, but are not limited to, complex branching morphology and multiple channel types.
\texttt{CARLsim}, \texttt{NEST}, \texttt{Brian2} and \texttt{BrainPy} focus on the dynamics and structure of neural
systems rather than on the detailed morphological and biophysical properties of individual neurons.
They can be used for simulating large heterogeneous networks.
A major advantage of \texttt{NEURON}, \texttt{NEST}, \texttt{Brian2} and \texttt{BrainPy} is that, in addition to the
built-in neurons and connection objects, users can also use low-level language (such as
C++) or mathematical model descriptions to design the dynamics of neurons and connections.
This provides convenience to investigate new mechanisms.
However, esoteric syntax may lead to a steep learning curve for new users.
In addition, the lack of automatic differentiation support makes these tools unsuitable for training SNNs for machine
learning tasks.

\subsection{Neuromorphic Computing Frameworks}
Neuromorphic Computing aims to apply the insights from neuroscience to create brain-inspired model and device.
SNN can be regarded as a core technique of neuromorphic computing.
Frameworks such as \texttt{Nengo} \cite{bekolay2014nengo}, \texttt{BindsNet} \cite{hazan2018bindsnet} and
\texttt{SpikingJelly} \cite{SpikingJelly} focus on behaviors of SNNs and can be applicated in the field
of machine learning.
They are built on deep learning framework (such as \texttt{PyTorch} and \texttt{TensorFlow}) to facilitate the fast
simulation of SNNs on CPU and GPU computational platforms, as well as using deep learning training procedures to
optimize model parameters.
\texttt{Nengo} is regarded as a cognitive modeling tool.
It can be used to build large-scale models based on the Neural Engineering Framework (NEF) to simulate advanced
functions of the brain or brain regions.
It also provides an extended version based on deep learning library, namely \texttt{NengoDL}, which uses
\texttt{TensorFlow} to improve simulation speed of \texttt{Nengo} models and implements automatic conversion
from \texttt{Keras} models to \texttt{Nengo} networks.
\texttt{SpikingJelly} and \texttt{BindsNet} are SNN libraries developed on top of the \texttt{PyTorch} deep learning
framework, which enable rapid prototyping and features concise syntax.
In general, they are closer to deep learning procedure and do not support neuroscience well, for example, simulating
the anatomical and biophysical properties of neurons and neural circuits.

These frameworks focus on either neuroscience or deep learning, rather than both.
\texttt{SPAIC} uses \texttt{PyTorch} as its matrix computations backend to perform efficient, which is suitable for
machine learning tasks.
In addition, several popular neuroscience-based neuron types, synapse types and connection types are provided for
users to choose from.
In this way, \texttt{SPAIC} can be seen as a bridge between the artificial intelligence computing and neuroscience
domains, enabling researchers to easily integrate neural mechanisms found in neuroscience with machine learning
techniques to develop better artificial intelligent.

\begin{table}[!htbp]
\caption{Comparison of neural network simulation tools regarding supported features.
An `\checkmark` denotes that the feature is supported by the simulator, and a blank denotes that the feature is not supported.}\label{FrameworkComparison}
\setlength{\tabcolsep}{1.1mm}{
\begin{tabular}{l c c c c c c c c c c c c}

\hline
& \makecell[c]{Tensor-\\Flow} & PyTorch & NEURON & GENSIS & \makecell[c]{CARL-\\sim} & NEST & Brian2 & BrainPy & Nengo & \makecell[c]{Binds-\\NET} & \makecell[c]{Spiking-\\Jelly} &
\textbf{SPAIC}\\
\hline
\multicolumn{13}{c}{Neuron model}\\
LIF& & &\checkmark& &\checkmark&\checkmark&\checkmark&\checkmark&\checkmark&\checkmark&\checkmark&\checkmark \\
aEIF& & & & & & \checkmark & \checkmark &\checkmark& & & &\checkmark \\
IZH & & & & &\checkmark&\checkmark&\checkmark&\checkmark&\checkmark& & &\checkmark \\
HH& & &\checkmark&\checkmark& &\checkmark&\checkmark&\checkmark&\checkmark&\checkmark& &\checkmark \\
\hline
\multicolumn{13}{c}{Synaptic type}\\
Chemical& & &\checkmark&\checkmark&\checkmark&\checkmark&\checkmark&\checkmark&\checkmark&\checkmark&\checkmark
&\checkmark \\
Electrical& & &\checkmark& &\checkmark&\checkmark&\checkmark&\checkmark&\checkmark& & & \checkmark \\
\hline
\multicolumn{13}{c}{Connection property}\\
Convolution& & & & & & & & & & \checkmark & \checkmark&\checkmark \\
Sparse&\checkmark&\checkmark&\checkmark& & & \checkmark& \checkmark&\checkmark & &\checkmark& &\checkmark \\
Loop& & &\checkmark&\checkmark&\checkmark&\checkmark&\checkmark&\checkmark&\checkmark& &
&\checkmark \\
Delay& & &\checkmark&\checkmark&\checkmark&\checkmark&\checkmark&\checkmark&\checkmark& & &\checkmark \\
\hline
\multicolumn{13}{c}{Learning algorithm}\\
STDP& & &\checkmark& &\checkmark&\checkmark&\checkmark&\checkmark&\checkmark&\checkmark& &\checkmark \\
Gradient&\checkmark&\checkmark& & & & & & & &\checkmark&\checkmark&\checkmark \\
\hline
\multicolumn{13}{c}{Computing hardware}\\
CPU&\checkmark&\checkmark&\checkmark&\checkmark&\checkmark&\checkmark&\checkmark&\checkmark&\checkmark&\checkmark
&\checkmark&\checkmark \\
GPU&\checkmark&\checkmark& & &\checkmark& & &\checkmark&\checkmark&\checkmark&\checkmark&\checkmark \\
\hline
\end{tabular}
}
\end{table}

\section{Package Structure}
\begin{figure*}[htbp]
	\centering
		\includegraphics[scale=.65]{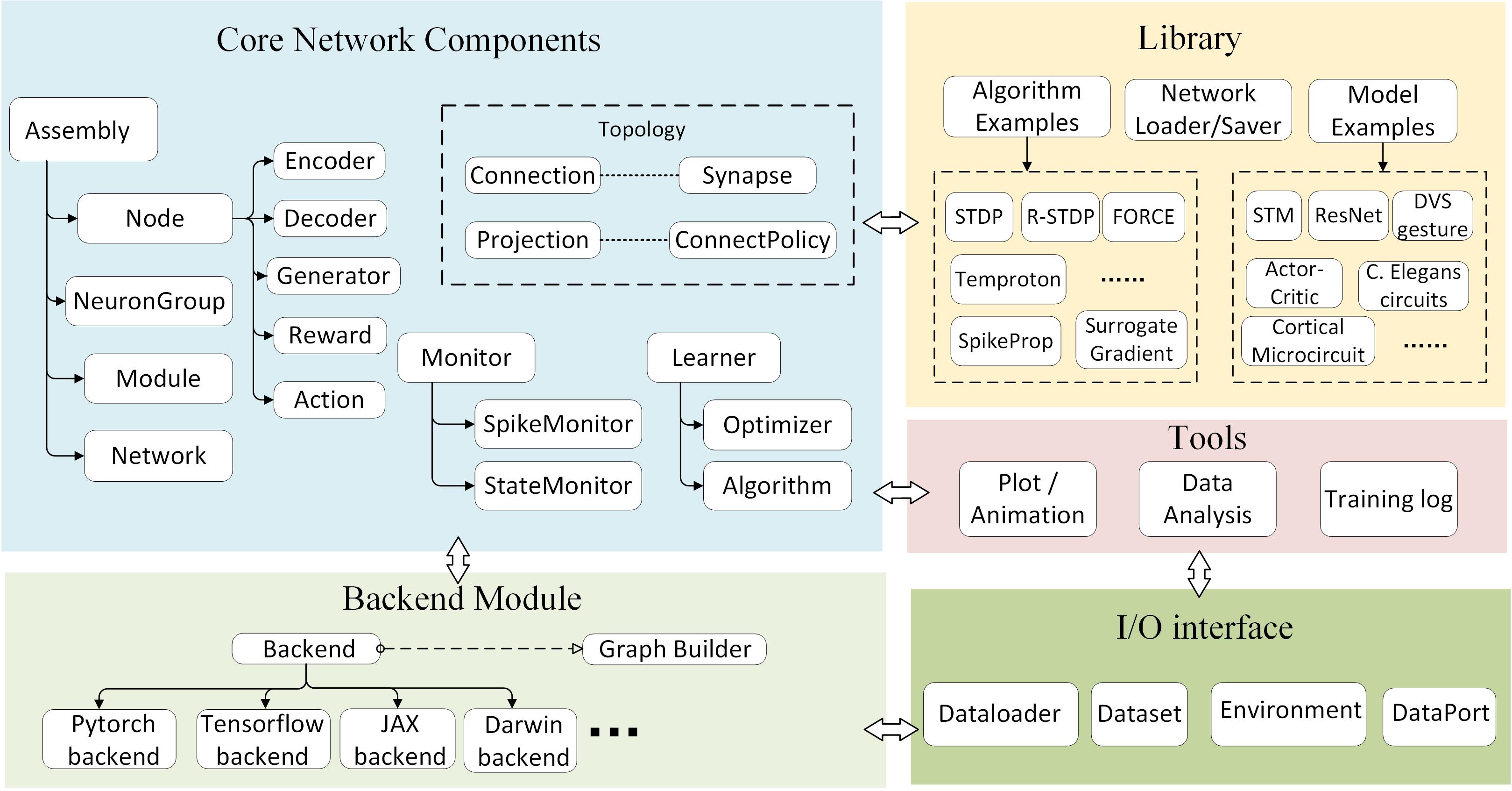}
	\caption{The structure of SPAIC}
	\label{Module_Structure}
\end{figure*}

The main structure of \texttt{SPAIC} is shown in Figure \ref{Module_Structure}.
The classes of \texttt{SPAIC} can be divided into three functional blocks.
\begin{enumerate}
    \item Network components contain all the components of the model which provide a frontend to set up network.
    \item Within simulation procedure, the simulation process of network is complied to computation graph in the \textit{Backend}.
    \item The training and analysis tools provide I/O interface, \textit{Plot}, \textit{TrainingLog} and et al.
\end{enumerate}

\texttt{SPAIC} provides a \textit{Network} object to contain all the network components.
Users can organize basic components, such as \textit{Nodes}, \textit{NeuronGroups}, \textit{Connections} and \textit{Assemblies},
with complex structure.
Auxiliary components, such as \textit{Learner}, \textit{Optimizer} and \textit{Monitor}, can be added according to
users' requirements.
A \textit{Backend} should also be attached to \textit{Network} to compile the frontend network model.

    \subsection{Network Components}
        \subsubsection{Assembly}
            First of all, \textit{Assembly} is one of the most important components of \texttt{SPAIC}.
            It is an abstract class of neural population, that \textit{Network}, \textit{NeuronGroup}, \textit{Node} and \textit{Module} are all inherited from it.
            It defines the basic network structure and attributes.
When users want to build a network with large scale and complex structure, especially models involving
multiple brain regions, they can construct \textit{Assembly} object with \textit{NeuronGroups} and
\textit{Connections} as the subregion. 
		The example code is shown as follow:
    \begin{lstlisting}
	# Construct an Assembly object as a subregion
	class SubRegionA(spaic.Assembly):
		def __init__(self):
			super(SubRegionA, self).__init__()

			self.l1 = spaic.NeuronGroup(10, neuron_model='lif')
			self.l2 = spaic.NeuronGroup(20, neuron_model='lif')
			self.connection1 = spaic.Connection(self.l1, self.l2)

	class TestNet(spaic.Network):
		def __init__(self):
			super(TestNet, self).__init__()

			...
			self.regionA = SubRegionA()
			...

		\end{lstlisting}

        \subsubsection{NeuronGroup}
            A \textit{NeuronGroup} is a group of neurons with the same neuron model and connection pattern.
            Also, a \textit{NeuronGroup} should contain all the details of neurons like the initial voltage and threshold.
            Users should provide model type and number of neurons when creating a \textit{NeuronGroup}.
\texttt{SPAIC} provides a series of build-in neuron models like leaky integrate-and-fire (LIF) model\cite{lif},
adaptive exponential integrate-and-fire (AEIF) model\cite{aeif}, Izhikevich (IZH) model\cite{izhikevich}, Hodgkin-Huxley (HH) model\cite{hh} and et al.
            The following code shows how to create a group with 50 LIF neurons.
    		\begin{lstlisting}
	# NeuronGroup
	self.l1 = spaic.NeuronGroup(neuron_number=50, neuron_model='lif')
   		 \end{lstlisting}

		Attributes of the neuron model, such as time constant $\tau_m$, can be modified as keyword arguments of \textit{NeuronGroup}.
		If there are no attributes given, \texttt{SPAIC} will use the default parameter values of these neuron model.
        In addition, users can define auxiliary attributes such as spatial position and neuron type in the \textit{NeuronGroup},
        which might be useful in constructing complex networks.

        \subsubsection{Node}
            \textit{Node} is the input and output convert unit of neural network, it contains coding mechanism which will convert input to spikes or convert spikes to output.
             It has five different subclasses:
    \begin{itemize}
        \item \textit{Encoder}:
        Compared with the static numerical input of ANN, SNN uses spike trains as input, which is consistent
        with the way brain transmits information.
        The \textit{Encoder} implements the function of converting the input data into input spikes. 
        \texttt{SPAIC} provides some common encoding methods, such as \textit{PoissonEncoder} and \textit{LatencyEncoder}.
        As an example, the following code defines a \textit{PoissonEncoder} object which transforms the input into
        poisson spike trains.
\begin{lstlisting}
	# Encode input
	self.input = spaic.Encoder(num=node_num, coding_method='poisson')
\end{lstlisting}

        \item \textit{Decoder}:
        The main usage of \textit{Decoder} is to convert the output spikes or voltages to a numerical signal.
        \texttt{SPAIC} provides some common decoding methods, such as \textit{SpikeCounts} and \textit{FirstSpike}.
        For example, the following code defines a \textit{SpikeCounts} object to get the number of spikes of each output neuron.
\begin{lstlisting}
	# Decode output spikes of layer2
	self.output = spaic.Decoder(num=10,
		dec_target=self.layer2,
		coding_method='spike_counts',
		coding_var_name='O')
\end{lstlisting}
        \item \textit{Generator}.
        It is a special encoder that will generate spike trains or current without dataset.
        For example, in some computational neuroscience studies, users need special input like poisson spikes to model background cortical activities.
        To meet such requirement, some common pattern generators, such as \textit{PoissonGenerator} or \textit{ConstantCurrentGenerator}, are provided in \texttt{SPAIC}.
        \item \textit{Reward}:
        During the execution of a reinforcement learning task, \textit{Reward}, a different type of decoder, is needed to decode the activity
        of the target object according to the task purpose.
        For example, the \textit{GlobalReward} for classification task determines the predicted result according to the
        number of spikes or the maximum membrane potential.
        If the predict result is the same as the expected one, the positive reward will be returned.
        On the contrary, negative rewards will be returned.
        \item \textit{Action}:
        \textit{Action} is also a special decoder that will transform the output to an action.
        The main usage of \textit{Action} is to choose the next action according to the action selection mechanism of
        the target object during reinforcement learning tasks.
        For example, the \textit{PopulationRateAction}, it will take the label of the neuron group with largest spiking
        frequency as action.
	\end{itemize}

	 \subsubsection{Topology}
The topology component is used to specify interactions between \textit{Assembly} objects, it is consist of \textit{Connection},
\textit{Synapse}, \textit{Projection} and \textit{ConnectPolicy}.
\textit{Connection} is the most generic implementation of topology that connects elementary \textit{Assembly} objects,
such as \textit{Node} and \textit{NeuronGroup}.
\textit{Connection} can specify \textit{Synapse} to define how the pre-assembly affects the post-assembly.
\textit{Projection} is an abstract topology representing the communication between high level \textit{Assembly} objects
which contains multiple \textit{Connections}.
\textit{ConnectPolicy} defines a rule to generate specific \textit{Connections} in \textit{Projection}.

\textit{Connection and Synapse}:
The \textit{Connection} is aware of pre-assembly and post-assembly object, as well as a matrix of weights of connections
strengths.
\texttt{SPAIC} supports a lot of different connection forms like \textit{FullConnection}, \textit{RandomConnection},
\textit{SparseConnection} and \textit{ConvolutionConnection}.
Noticeable, the \textit{SparseConnection} is developed for the simulation of large scale network with sparse connectivity.
In addition, the synapse is a critical structure of neural connection that usually transmit information from the source
neurons to the target neurons.
The \textit{Synapse} object can be specified when creates a \textit{Connection} object and the input is filtered by the
synapse kernel.
If there is no specific \textit{Synapse} in a \textit{Connection} object, the weighted sum of input from pre-assembly is directly
added to the state variable of post-assembly.
For example, the following code shows how to build a full connection that connect input layer and layer1:
\begin{lstlisting}
	# Connection between input layer and layer1
	self.connection1 = spaic.Connection(
				pre_assembly=self.input,
				post_assembly=self.layer1,
				link_type='full')
\end{lstlisting}

\texttt{SPAIC} provides some classical \textit{Synapse} types in neuroscience, such as \textit{NMDASynapse}, \textit{AMPASynapse},
\textit{GABASynapse} and \textit{GapJunction}.
Delay is also important in neuroscience, but it is difficult to achieve on a general deep learning platform.
In \texttt{SPAIC}, delay can be added to connections with extra memory consumption.

\textit{Projection and ConnectPolicy}:
The \textit{Projection} is a high-level network component that can contain multiple \textit{Connections} between the
\textit{NeuronGroups} in pre-assembly and post-assembly.
It can automatically generate \textit{Connections} between \textit{NeuronGroups} according to user defined rules, which
is useful when constructing complex networks with repetitive connection patterns.
The \textit{ConnectPolicy} object provides the interface for users to define the connection rules, and be used in the
\textit{Projection} object.
\texttt{SPAIC} provides some prototype policies such as \textit{IncludeTypePolicy}, which is to connect the
user specified types (e.g. pyramidal, inhibitory) of \textit{NeuronGroups}.
In addition, users can define more complex connect behavior by overriding the \texttt{\_\_init\_\_}function of
the \textit{Projection} object.
For example, the following code shows how to generate connections between excitory neurons and inhibitory neurons in two
layers:
\begin{lstlisting}
	# Define Assembly with different types of neurons
	class SubRegion(spaic.Assembly):
		def __init__(self):
			super(SubRegionA, self).__init__()
			self.l1 = spaic.NeuronGroup(10, neuron_model='lif', type='excitory')
			self.l2 = spaic.NeuronGroup(20, neuron_model='lif', type='inhibitory')
		layer1 = SubRegion()
		layer2 = SubRegion()
	# Define specific connection rules and pass it to a Projection object
	ei_policy = spaic.IncludeTypePolicy(pre_types=['excitory'], post_types=['inhibitory'])
	ei_project = spaic.Projection(pre=layer1, post=layer2, policies=[ei_policy])
\end{lstlisting}

    \subsubsection{Module}
        The \textit{Module} is a special subclass of \textit{Assembly} which is directly inherited from \textit{torch.nn.Module}.
So \texttt{SPAIC} can implement functions supported by \texttt{PyTorch} and has advantage to combine ANN and SNN in a hybrid network.
The following example shows how to use \textit{Module} to add a convolution layer and a batch normalization layer into a network.

\begin{lstlisting}
	# Module with a convolution layer and a batch normalization layer
	self.layer1_layer2_con = spaic.Module(nn.Sequential(
            nn.Conv2d(
                in_channels=in_channel,
                out_channels=out_channel,
                kernel_size=3
            ),
            nn.BatchNorm2d(out_channel)
        ),
		input_targets=self.layer1, input_var_names='O',
		output_tragets=self.layer2, output_var_names='WgtSum')
\end{lstlisting}

	\subsubsection{Monitor}
\textit{Monitor} records the state variables of connections and neurons during whole simulation time.
At present, there are two types of monitors in \texttt{SPAIC}, the \textit{SpikeMoniter} and \textit{StateMonitor}.
The \textit{SpikeMoniter} records the spike trains and the \textit{StateMonitor} records all other states.
The following example code shows how to build a \textit{StateMonitor} and a \textit{SpikeMoniter} to record voltage
and output of layer1, respectively.
\begin{lstlisting}
	# Monitor the voltage and output of layer1
	self.mon_V = spaic.StateMonitor(self.layer1, 'V')
	self.mon_Spike = spaic.SpikeMonitor(self.layer1, 'O')
\end{lstlisting}

	 \subsubsection{Learner}
	 	\textit{Learner} is a base class for all learning and optimization algorithms of the network.
	 	Learning algorithm is the main part of learner that can access and update any specified network parameters.
        The exact learning algorithm can be defined by users as they need.
        Here, we provide two common types of algorithms of SNN as examples.
	 	The first type of algorithms is based on gradient back propagation like \texttt{STCA}\cite{stca} and \texttt{STBP}\cite{stbp},
        which mainly realized by surrogate gradient.
	 	Another kind is based on synaptic plasticity like \texttt{STDP}\cite{stdp} which is more biologically plausible.
	 	Meanwhile, \texttt{SPAIC} also has built in algorithms for reinforcement learning, like \texttt{RSTDP}\cite{rstdp},
        which is based on \texttt{STDP} learning rule with reward mechanism.

		\textit{Optimizer} is another part of learner.
		It contains many optimization algorithms like \textit{Adam} and \textit{SGD} which can be used to
        optimize network parameters of gradient based algorithms.
		The usage of gradient based learning algorithm \textit{STCA} and optimization algorithm \textit{Adam} is as follow:
\begin{lstlisting}
	# Learner
	self.learner = spaic.Learner(trainable=self,
		algorithm='STCA')
	self.learner.set_optimizer('Adam', 0.001)
	...
	learner.optim_zero_grad()
	learner.optim_step()
\end{lstlisting}

        \subsubsection{Network}
            \textit{Network} is the top-level of the model that all other components should be included in it.
The network model can be defined by inheriting the \textit{Network} class and adding network components in
the \textit{\_\_init\_\_} function.
We also provide two kinds of network construction methods which are described in detail in the document of \texttt{SPAIC}.
The \textit{run} function of \textit{Network} class starts the simulation of all network components in \textit{Backend}.
The following code shows how to construct and run a network model.

\begin{lstlisting}
	# Construct a TestNet
	class TestNet(spaic.Network):
	    def __init__(self):
	        super(TestNet, self).__init__()
	        ...

	net = TestNet()
	...
	net.run(run_time)
\end{lstlisting}

    \subsection{The Simulation Procedure}
\begin{figure}[htbp]
    \centering
        \includegraphics[scale=.60]{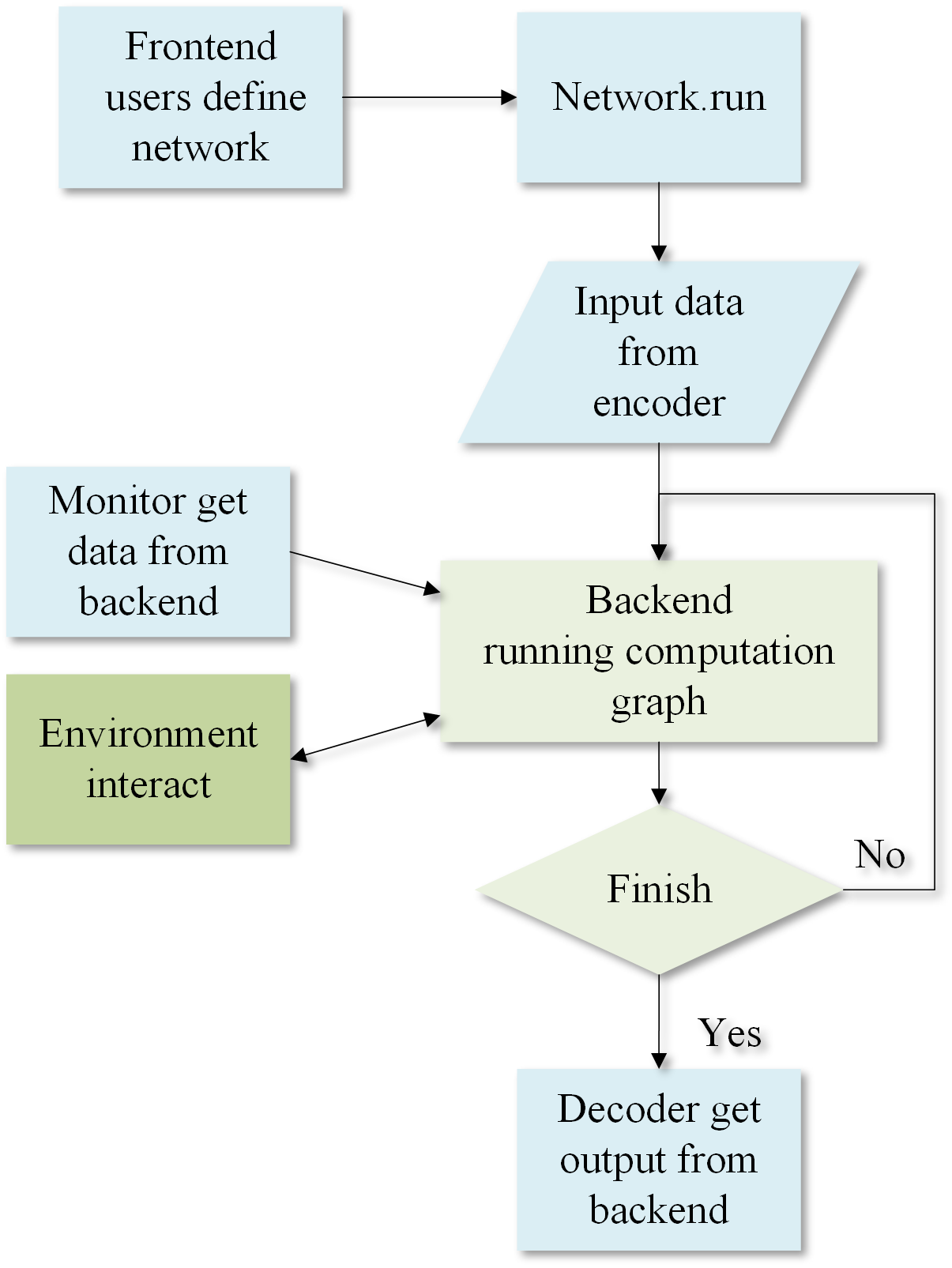}
    \caption{The computation flow in \texttt{SPAIC}.}
    \label{ComputationFlow}
\end{figure}

\texttt{SPAIC} decouples neural model creation and simulation into the frontend and the backend, respectively.
The network components described above provide the frontend interfaces to create static symbolic description of the network.
The \textit{Backend} can take the model described in the frontend, transform the neuron and synaptic models into discrete
equations, and build optimal computation graph that are suited to implement the simulation.
When the simulation starts, the \textit{Backend} fetches data from \textit{Encoder} and performs operations of the entire computation graph.
Meanwhile, the \textit{Decoder} or \textit{Monitor} can get data from \textit{Backend}.
The whole process is shown in Figure \ref{ComputationFlow}.

Building graph is a key step in the procedure from frontend network representation to backend simulation.
Biological neural circuits are cyclic graph, hence brain inspired model cannot be built directly due to circular
dependency on calculation.
\texttt{SPAIC} provides two build strategies which build model into parallel or serial computation graph respectively.
Firstly, the parallel computation graph is the most common way in neuroscience SNN simulators that all of the connections will use the
output of pre-assemblies from the last step so that all components can run in parallel.
Secondly, to be consistent with the serial computation in DNN, \texttt{SPAIC} provides another one build strategy.
It uses Topological Sorting\cite{topologicalsorting} to search the network and decompose the network structure into
feedforward and cyclic parts.
In the cyclic part, one step delay is added to one of connections to solve the circular dependency problem.
Then network model can be transformed into Directed Acyclic Graph (DAG).

It is noted that \texttt{SPAIC} supports both static and dynamic computation diagrams.
Although it mainly uses \texttt{PyTorch} as backend which uses dynamic diagram for computation, the architecture
of \texttt{SPAIC} is designed to generate a static computation when the build process is complete.
At the same time, the design of static diagram makes \texttt{SPAIC} support multiple backends, such as
\texttt{TensorFlow} and \texttt{Jax}.
On the other hand, we provide dynamic computation diagram that allows some operation modifications in
computation graph during runtime.

    \subsection{Training and Analysis Tools}
    		Training and analysis are the most time-consuming works of neural network research.
 			\texttt{SPAIC} supports a number of useful tools and I/O interface to provide a more user-friendly environment.
 			\textit{Plot} and \textit{TrainingLog} are such major tools: \textit{Plot} object provides functions for generating diagram of results and process analysis, and \textit{TrainingLog} will record the variation of parameters to help users analyze the role of algorithms.

 			The I/O interface component provides four tools: \textit{DataLoader}, \textit{Dataset}, \textit{DataPort} and \textit{Environment}.
    		\begin{itemize}
    			\item \textit{DataLoader}: organizes and format data, such as generate index from data and
                batch it according to the batch size.
    			\item \textit{Dataset}: loads data from dataset according to the format of storage.
    			\item \textit{DataPort}: a data transmission interface that can receive data in real time.
    			\item \textit{Environment}: used for reinforcement learning.
    		\end{itemize}

    		Finally, \texttt{SPAIC} contains \textit{NetworkSave} component in library tools, which let users can save the parameters or the whole trained model for further use.

    \subsection{Custom Extension}
    		\texttt{SPAIC} can also be extended by custom-defined neuron models, connections
and learning rules with customization mechanisms,
which we will discuss in the following.
\subsubsection{Customization mechanisms}
We customize network components through adding variables and operations.
There are two types of coding methods as below.
	\begin{itemize}
        \item \textit{String based command}:
        Objects of \texttt{SPAIC} (e.g., \textit{NeuronGroup}, \textit{Connection}, etc.) provide a
        variables dictionary and an operations list.
        When users want to implement an operation like A=B+C.
        The first step is to add variables used in the operation into variables dictionary as below.
\begin{lstlisting}
  self._variables['A']=0.0
  self._variables['B']=0.0
  self._variables['C']=0.0
\end{lstlisting}
        The second step is to append the string of the operation into operations list.
        The main format of operations is that the first variable represents the result, the second
        item is the name of the basic operations supported in \textit{Backend}.
		The remain variables from the third term represent the input parameters of the calculation
        formula.
\begin{lstlisting}
	self._operations.append(('A', 'add', 'B', 'C'))
\end{lstlisting}

   \item \textit{Standalone function}:
When the calculation formula is hard to be implemented using basic operations provided in \textit{Backend},
users can use the standalone function approach.
Firstly, the desired operation needs to be implemented in a Python function as below.
\begin{lstlisting}
	def add_func(a, b):
        return a+b
\end{lstlisting}
Then, the string of the operation should be appended into operations list,
where the second item is the callable function.
\begin{lstlisting}
	self._operations.append(('A', add_func, 'B', 'C'))
\end{lstlisting}
    \end{itemize}

\subsubsection{Custom Neuron Model}
The abstract class \textit{NeuronModel} implements functionality that is common to all neuron types.
The user must define the calculation formula of the neuron themselves in the body of the \textit{\_\_init\_\_()}
function.
Then the customized model should be registered to the \texttt{SPAIC} using the \textit{NeuronModel.register} function.

\subsubsection{Connection}
Users of \texttt{SPAIC} can define their own connection types by creating a class that inherits from
\textit{Connection}.
To define a new connection class, one must override the \textit{unit\_connection} function.

There are many different synapses in the neural systems, \texttt{SPAIC} allows users to add their own synapse models.
\texttt{SPAIC} provides abstract class \textit{SynapseModel} which is similar with \textit{NeuronModel}.

\subsubsection{Learning rule}
\textit{Learner} is an abstract class for all learning and optimization algorithms of the network.
New learning algorithm class should override \textit{custom\_rule} function of \textit{Learner} to access
and update any specified network parameters.
We define trainable list and pathway list.
The trainable list contains the network components whose parameters can be updated by the learning rule.
The pathway list contains the network components are useful to the learning rule but whose parameters should not be updated.
For example, in a gradient back-propagation learning rule, if only last few layers of network should be trained, the rest of
layers should be added to pathway list so that the loss gradients can be passed through the whole network.

    There are two families of learning rules supported in \texttt{SPAIC}, gradient based or plasticity based.
    Users can follow the format of \texttt{STCA} or \texttt{STBP} in \texttt{SPAIC} to customize gradient based
algorithm.
    If users want to add plasticity based algorithms, they can follow the format of \texttt{STDP}.

\section{Examples}

    \subsection{Computational Neuroscience experiment}
The central motivation of \texttt{SPAIC} is to assist the development of intelligence computing models inspired by the
biological brain.
Hence, it is fundamental that \texttt{SPAIC} can support models and methods used in computational neuroscience, and then
introduce those features into a brain-inspired AI system.

\subsubsection*{Case study 1: cortical microcircuit model}
\begin{figure}[!htb]
    \centering
    \includegraphics[scale=0.3]{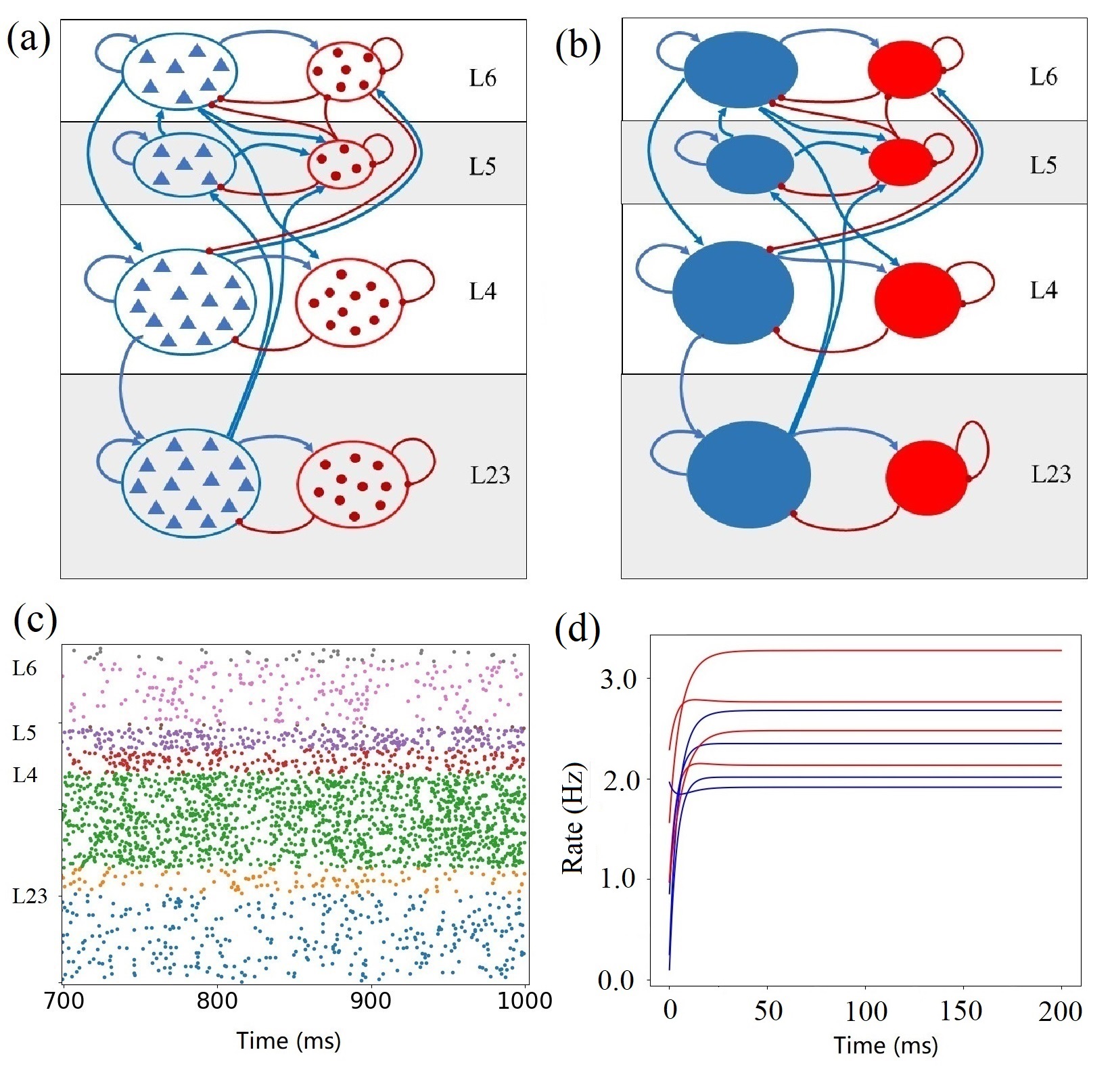}
    \caption{Cortical microcircuit network structure and activity. (a) Microcircuit structure of the spiking
    network model. (b) Microcircuit structure of the mean-field model. (c) Activity of the spiking network model.
        (d) Activity of the mean-filed model. }
    \label{Cortical}
\end{figure}

We first demonstrate the implementation of a spiking cortical microcircuit model using \texttt{SPAIC}, and use a
mean-field counterpart model to show the multi-scale modeling in \texttt{SPAIC}.
As shown in Figure \ref{Cortical}(a), the Potjans-Diesmann model contains two types of neuron distributed in four layers
[L23, L4, L5, L6], which represents cortical microcircuit network below a surface of 1 mm \cite{potjans2014cell}.
In each layer, the subnetwork can be viewed as an internally connected excitation and inhibition balance
network \cite{einetework}, which can be built as a prototype submodule using the \textit{Assembly} object.
Then the cortical microcircuit can be constructed by stacking and connecting the four layers with different parameters.
The realization code of above network is shown in Figure \ref{CorticalCode}, here we build neuron groups with LIF neuron model,
use \textit{SparseConnection} to randomly connect neurons with a given probability, and use \textit{PoissonGenerator}
to model external noise input to each neural population.
As shown in Figure \ref{Cortical}(c), the model exhibits irregular and stationary spiking activities similar with the
result in original work.
In addition, we can replace the LIF spiking model with a neural mass rate model and connect \textit{NeuronGroups} with
trainable coupling efficiency, then we can train this abstract model to have equivalent result (Figure \ref{Cortical}
(b)(d)).

\subsubsection*{Case study 2: C. elegans thermotaxis circuit}
\begin{figure}[htbp]
    \centering
        \includegraphics[scale=.5]{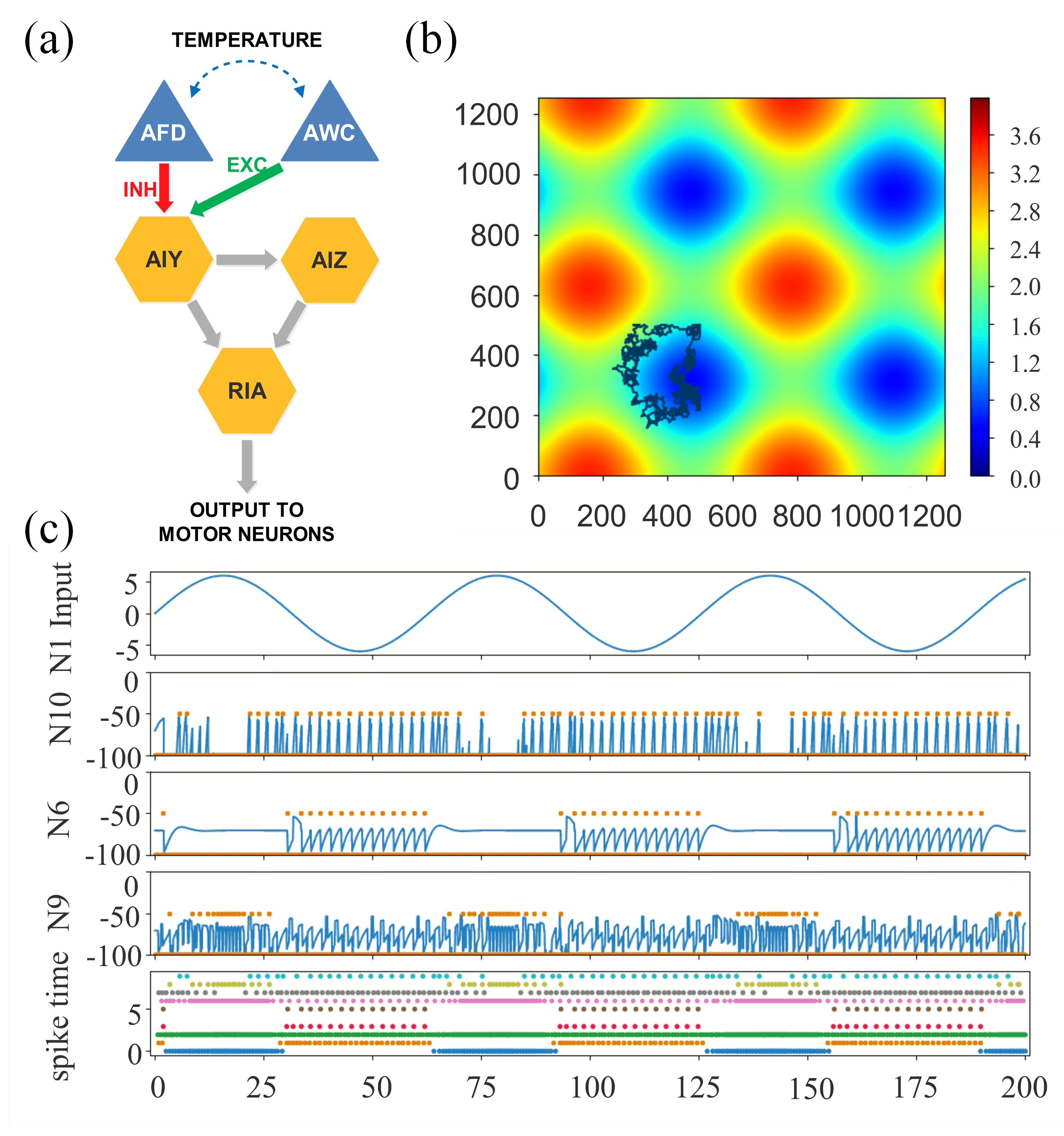}
    \caption{C. elegans temperature sensation modeling and simulation.
        (a) C. elegans thermotaxis circuit.
        (b) Simulation of C. elegans thermotaxis based contour tracking. (c) Activity of the spiking network model.}
    \label{ElegansNetwork}
\end{figure}

The second example of \texttt{SPAIC} application is a spiking model of C. elegans thermotaxis circuit, adapted from
earlier studies \cite{kimata2012thermotaxis, bora2014mimicking}.
This network has a small number of well-characterized neuron types (Figure \ref{ElegansNetwork}(a)) and is known to generate a
stereotypical triphasic motor pattern (turn clockwise, turn anticlockwise and random walk) when sensor neurons receive
temperature changes.
The design of interneurons structure performs as the derivative operation when the opposite directions of currents get
through different length of trails, and thus supports gradient detector.
This simple mechanism allows C. elegans to achieve diversion and track the favorable temperature contour.
The simulation of the network in \texttt{SPAIC} is shown in Figure \ref{ElegansNetwork}(c), we use aEIF model for
modeling the dynamics of the neurons, adopt \textit{full Connection} to link single neurons.
As the input temperature changes above or below the threshold, the output neurons of turn clockwise or anticlockwise
will spike alternately.
Thus it can control the temperature steering of C. elegans and make it track along the optimum temperature region as
Figure \ref{ElegansNetwork}(b) demonstrated.

\subsection{Training Deep Spiking Neural Networks}
\texttt{SPAIC} is suitable for training deep SNNs to solve problems in the domain of machine learning.
Here, we present example scripts to show how to build deep SNNs using \texttt{SPAIC} to implement machine
learning tasks, such as speech recognition and image recognition.
\subsubsection*{Case study 3: speech recognition}
\begin{figure*}[htbp]
    \centering
    \includegraphics[width=.8\linewidth]{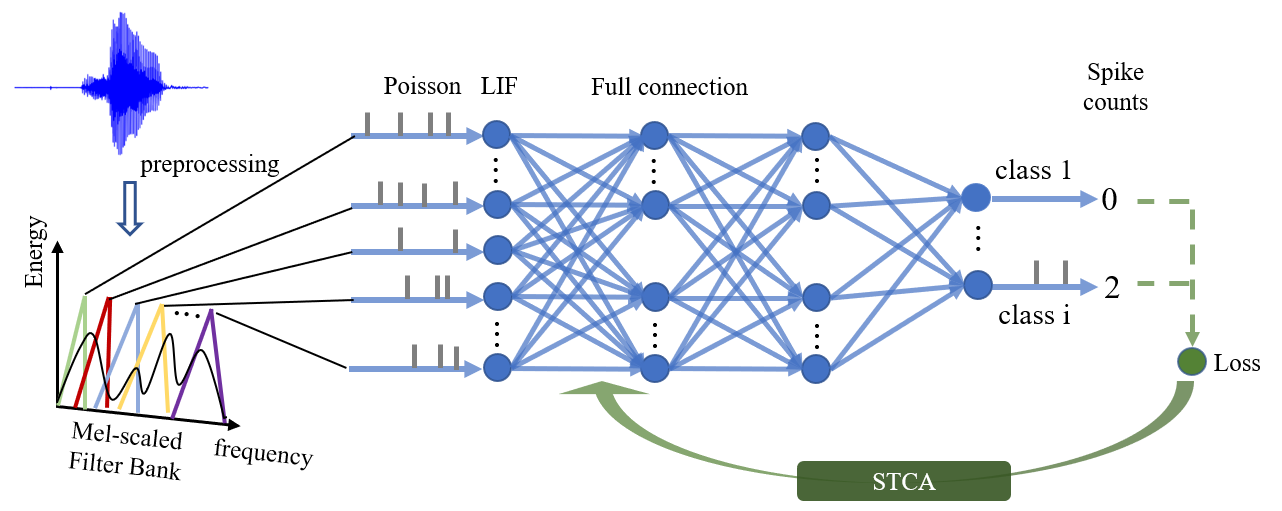}
    \caption{The network structure for TIDIGITS speech recognition.
    It is a four-layer SNN with full connection which is trained by \texttt{STCA} algorithm.}
    \label{speech_structure}
\end{figure*}

As shown in Figure \ref{speech_structure}, we trained a four-layer SNN to implement supervised learning
of the TIDIGITS speech dataset \cite{leonard1993tidigits}, which consists of digit utterances from `zero' to `nine' and `oh'.
The dataset is split into 3465 training samples and 1485 testing samples.
It is worth noting that when reading speech files directly, there will be tens of thousands of data in one second of
audio with a lot of redundant information.
Therefore, before performing the speech recognition task, we need to preprocess the raw speech files to obtain the
features of the dataset.
In \texttt{SPAIC}, we provide two popular preprocess methods, namely, Mel-frequency
cepstral coefficients (MFCC) \cite{usman2019dataset} and keypoints extraction (KP) \cite{xiao2018spike}.
The code of implementing SNN in \texttt{SPAIC} to classify TIDIGITS is given in Figure \ref{speech_code}.
Here, we use \textit{MFCC} preprocess method to extract features and use \textit{STCA} learning algorithm to train entire network.

\begin{figure}[htbp]
    \centering
    \includegraphics[width=.7\linewidth]{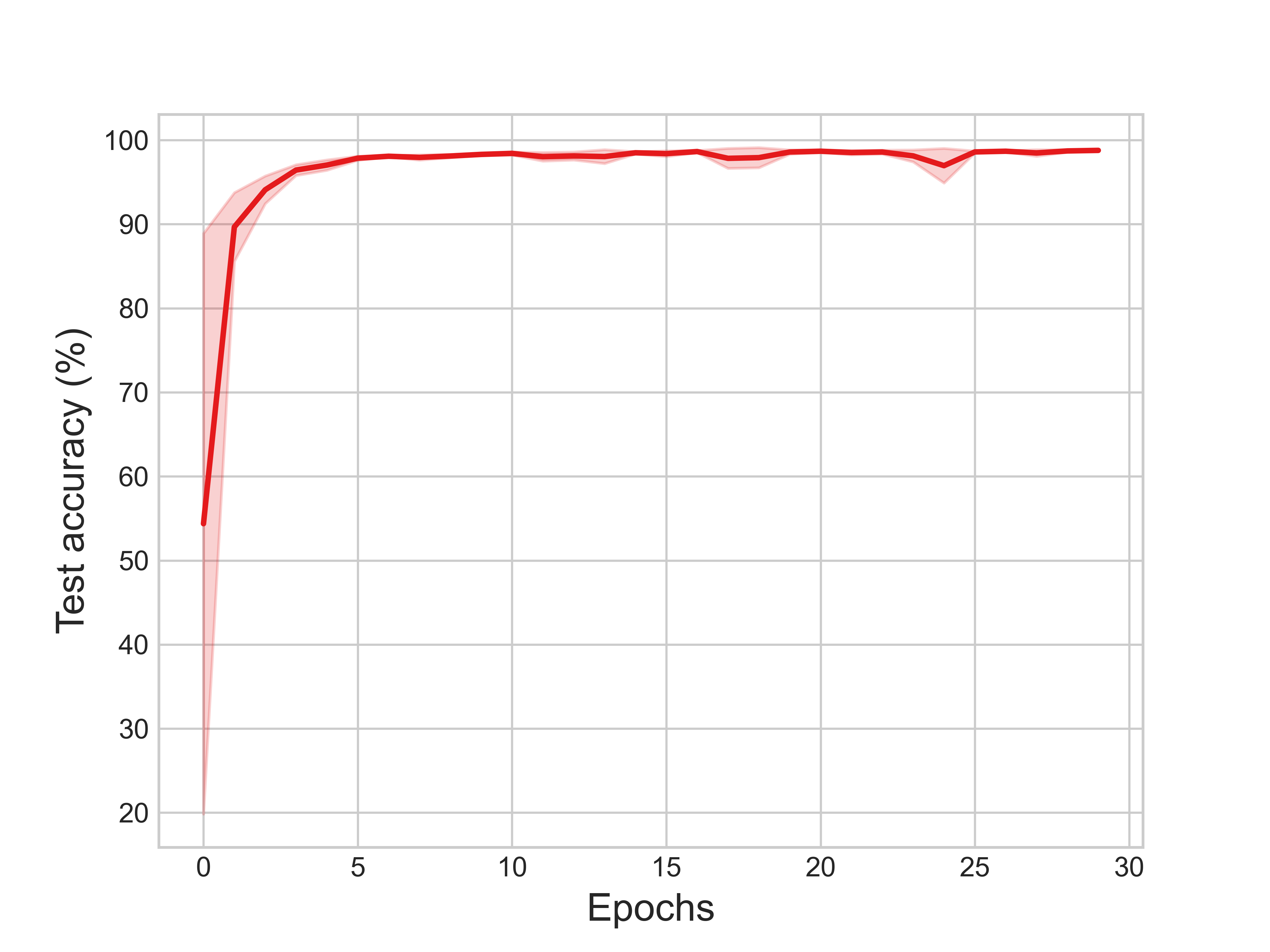}
    \caption{The network can developed suitable representations as demonstrated by increased average test accuracy
    over 5 runs.}
    \label{speech_result}
\end{figure}

The results in Figure \ref{speech_result} show that competitive performance has been achieved within 5 training epochs.
These results demonstrate the effectiveness of the preprocessing and network training method,
making \texttt{SPAIC} suitable for training deep SNNs to solve machine learning tasks.

\subsubsection*{Case study 4: image recognition}
\begin{figure}[htbp]
    \centering
    \includegraphics[width=.6\linewidth]{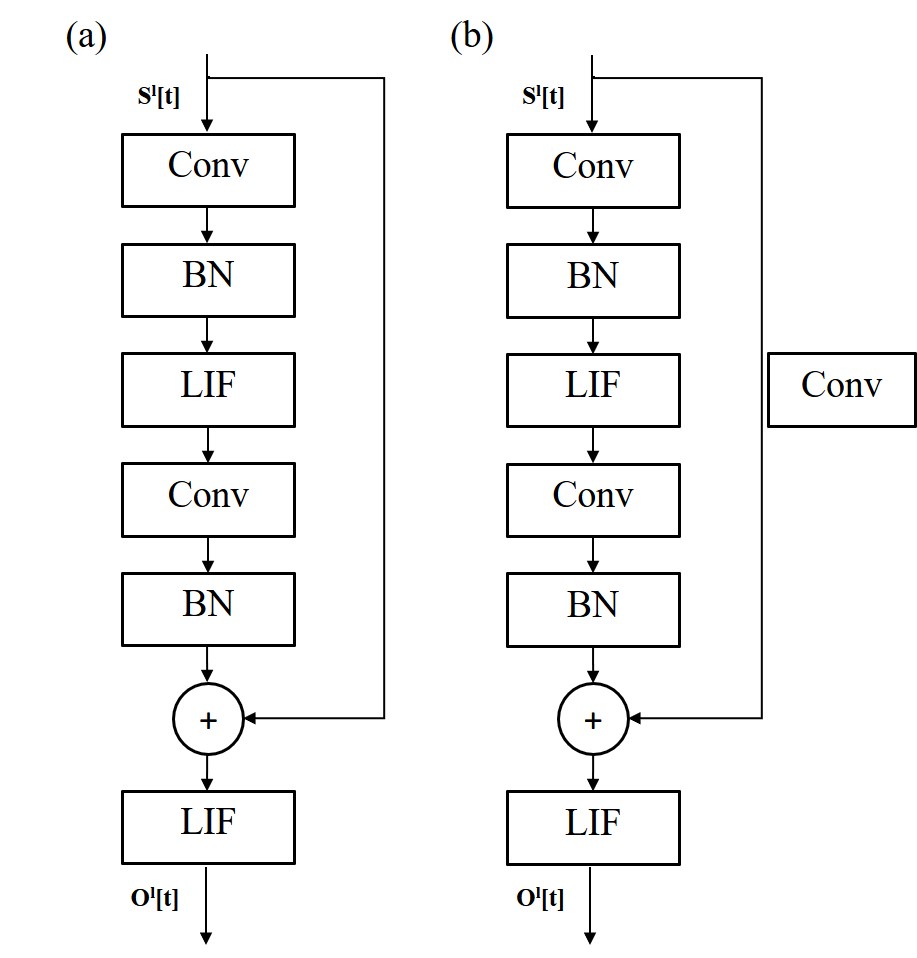}
    \caption{The structure of residual block. (a) Basic block. (b) Residual block with downsample.
    If the input and output have different dimensions or $stride$\textgreater1, convolution will be performed on the
    shortcut.}
    \label{residual_structure}
\end{figure}

In addition to constructing SNN with fewer layers, \texttt{SPAIC} is also convenient for implementing deeper
SNN.
In recent years, deep residual network (ResNet) has replaced VGG as the basic feature extraction network in
computer vision \cite{he2016deep}.
Here, we implement a spiking ResNet-18 on \texttt{SPAIC}.
The spiking residual block is the key structure to implementing ResNet.
When the residual block is built, the ResNet with arbitrary depth can be realized through repeated structure.
In spiking residual block, we replace ReLU activation layers of standard residual block in ANNs with spiking neurons.
The basic structure of spiking residual block is shown in Figure \ref{residual_structure}(a).
For basic block, the input and output have the same dimension, so there is no convolution operation on the shortcut
path.
When the input and output have different dimensions or $stride$\textgreater1, there is a convolution operation on the
shortcut, and the structure of residual block is shown in Figure \ref{residual_structure}(b).
The example code of implementing residual block is given in Figure \ref{image_code}.
Using \textit{Assembly} and \textit{Module} classes of \texttt{SPAIC}, we can conveniently implement residual block,
or any complex deep model structure we want.

\subsection{Neuromorphic Computing Applications}
Another important use case of \texttt{SPAIC} is to support neuromorphic computing applications that will eventually
perform on robots or terminal systems.
These situations require the platform to 1) run fast enough to be able to response in real-time and 2) provide data
interfaces to communicate with the other system during the network simulation.
Here, we present a simulated robot implementing reinforce learning task to show the usability of \texttt{SPAIC} in
those applications.

\subsubsection*{Case study 5: SNN and ANN hybrid reinforcement learning}
\begin{figure}[htbp]
        \centering
            \includegraphics[scale=.60]{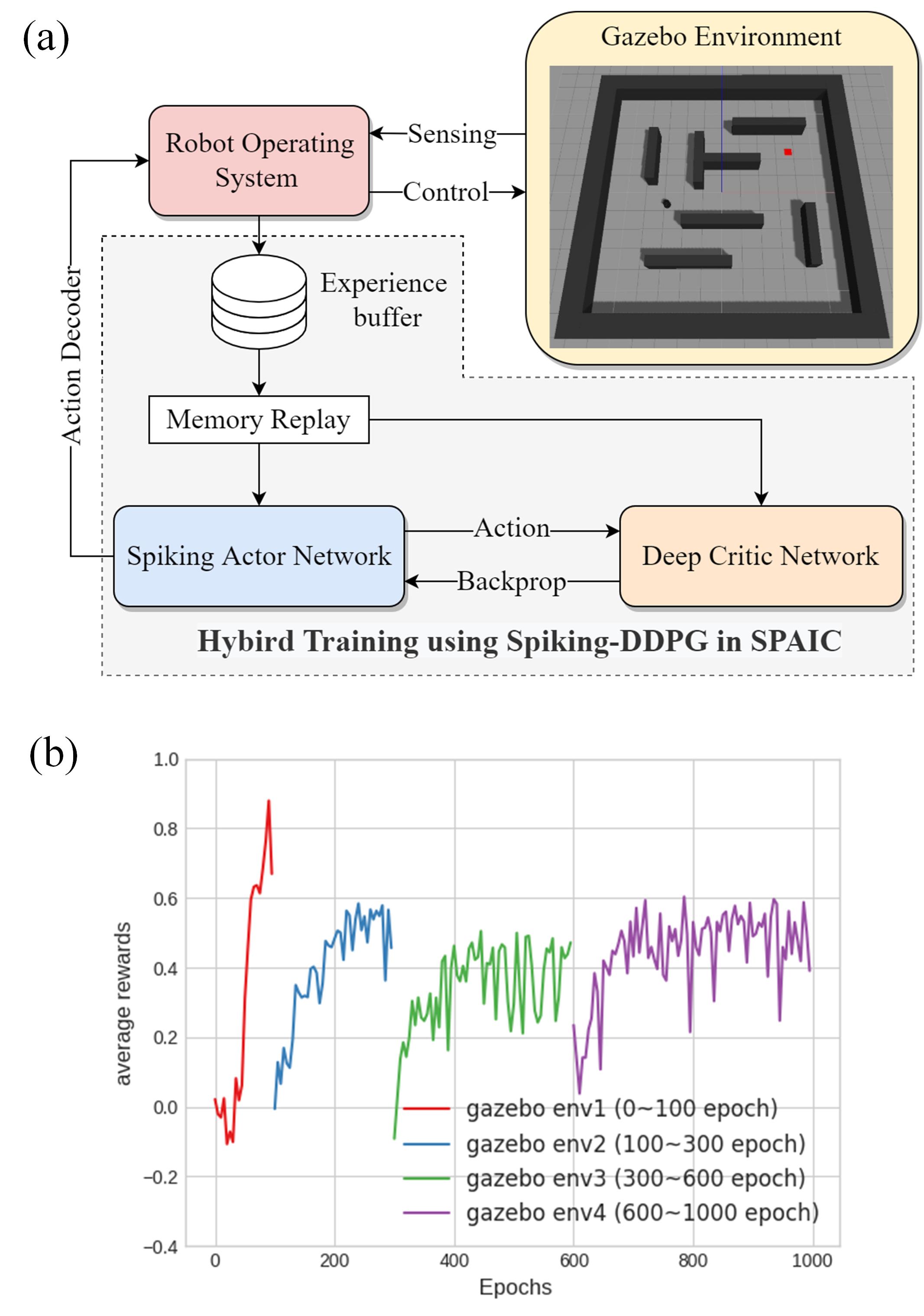}
        \caption{Training Spiking-DDPG on the SPAIC framework. (a) The structure of Spiking-DDPG training system. (b)
            The result of training.}
        \label{SpikeDDPG}
    \end{figure}

Reinforcement learning is other kind of important machine leaning algorithm, which is concerned with learning to control an
agent (such as robots) to maximize the performance in respect to a long-term objective.
As an example, we adopt the Reinforcement co-Learning method(Spiking-DDPG), which consists of a spiking actor network(SAN) and a deep critic
network, to construct an end-to-end SNN network for mapless navigation task of the wheel robot (turtlebot2).
The Gazebo Robot simulator is used as a middleware for both the training and validation.
\textit{PoissonEncoder} is used to encode the input state obtained from the Gazebo simulator, which is a 1x24
array composed of the relative distance and direction from the robot to the goal, the linear and angular velocities
of the robot, and the distance observations from the laser range scanner.
During training, the SAN built in \texttt{SPAIC} generated an action for the input state given by the Robot simulator.
Then, the action will be fed to the deep critic network for predicting the associated Q-value for the training of SAN.
Meanwhile, the action which takes spiking rates of 2 neurons will be transformed to the left and right wheel speeds of the
differential drive robot with the \textit{SpikeCount} Decoder.
Then, it will be published to the Gazebo environment for generating a reward to update the action-value for
the training of the deep critic network.
The \texttt{SPAIC} platform supports a wide range of neurons and training policies which can be
applied by users to modify the SNN actor network to perform any end-to-end robot control tasks.
Figure \ref{SpikeDDPG} depicts the training and validation environment instance in Gazebo simulator and
the details of the implementation of Spiking-DDPG training work.

    \subsection{Benchmark}
    To compare with the existing prominent platforms, we test the performance of \texttt{SPAIC} by running a simple network with
increasing neuron scales.
We simulated a network with one input layer of 100 neurons and Poisson input of 30HZ.
The core test layer of this network has $n$ (from 10 thousand to 10 million) LIF neurons to test the running speed.
We run each test model for 100ms with time step $dt=1.0ms$.
The model will be run 100 cycles and the whole running time is computed.
All benchmark tests run on the workstation with Ubuntu 18.04 LTS and Intel(R) Xeon(R) Gold 6230 CPU @ 2.10GHz, 128Gb RAM and a Nvidia Quadro GV100 GPU with 32 GB DRAM.
Python 3.8 is used by all tests.

	We tested \texttt{SPAIC}, \texttt{SpikingJelly}, \texttt{BindsNet} and \texttt{BrainPy} with both CPU and GPU, and
tested \texttt{Brian2} with only CPU.
	\texttt{SpikingJelly} and \texttt{BindsNet} are based on \texttt{PyTorch} and can support training of surrogate gradient
algorithms, but \texttt{SpikingJelly} does not have the mechanism of synaptic plasticity.
	\texttt{BrainPy} and \texttt{Brian2} are focus on neural dynamics simulation that cannot train or
don't have gradient back propagation mechanism.
 To provide an impartial comparison, we choose to simulate the network without training.
We have considered \texttt{Neuron}, \texttt{GENSIS}, \texttt{CARLsim}, \texttt{NEST} and \texttt{Nengo}, but in these cases, we were unable to implement the benchmarked network structure or the performance is significantly low in such benchmark.

\begin{figure*}[!t]
\centering
\subfloat[]{\includegraphics[width=2.5in]{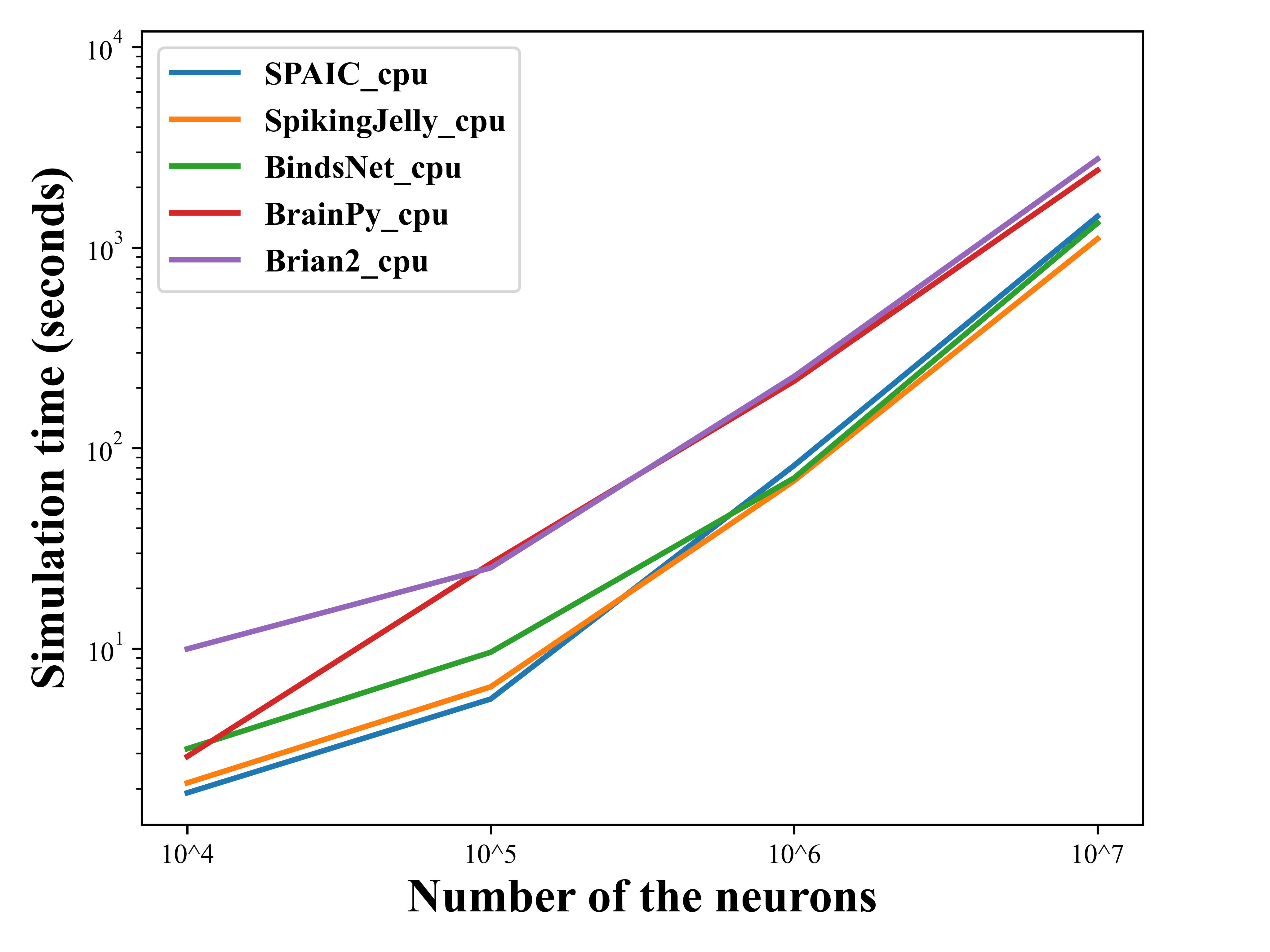}
\label{cpu benchmark}}
\hfil
\subfloat[]{\includegraphics[width=2.5in]{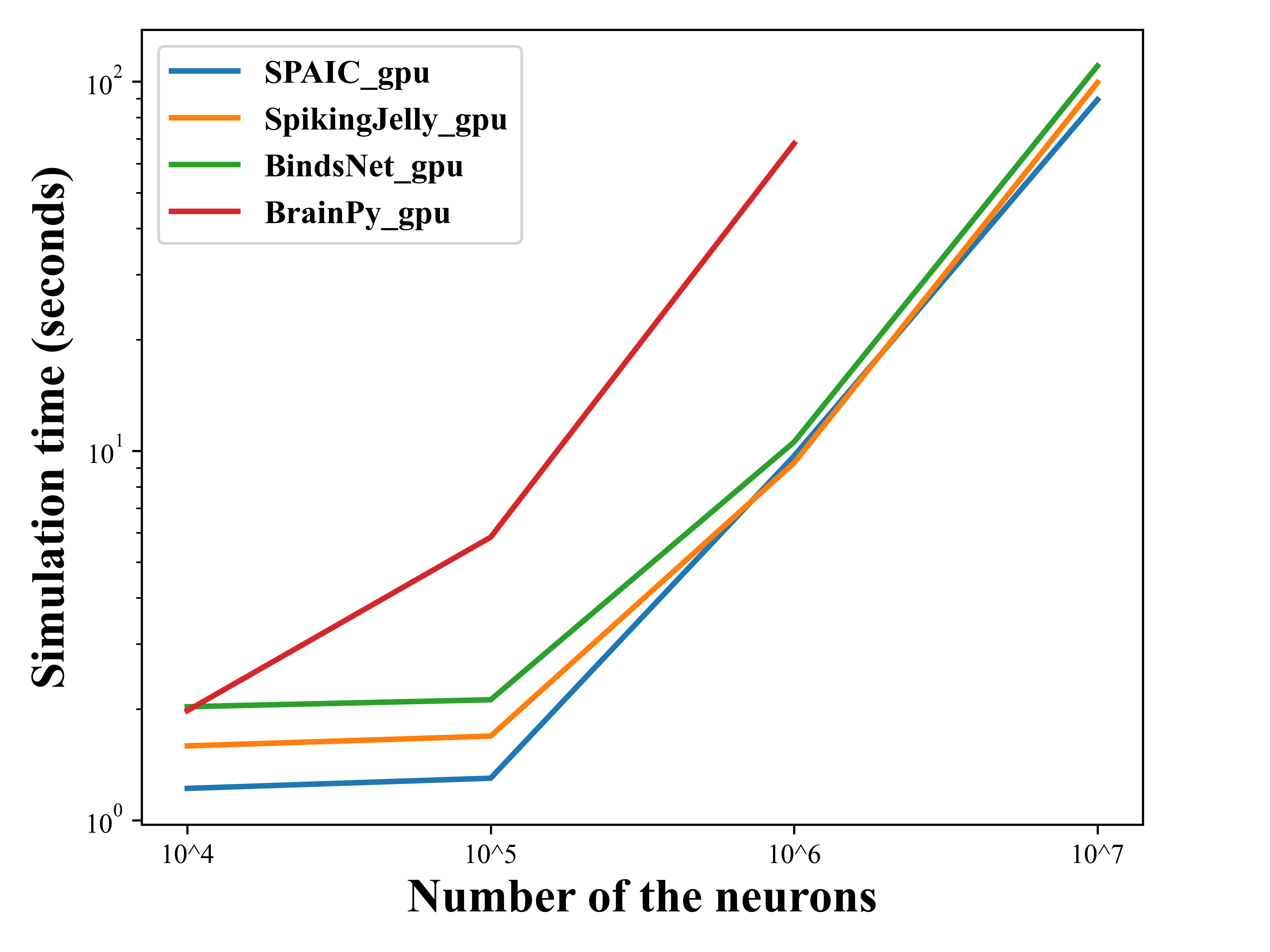}
\label{gpu benchmark}}

\caption{The benchmark result. (a) Time cost of benchmark test in CPU device. (b) Time cost of benchmark test in GPU device. note: performance of Brian2-GPU is not considered since its support of GPU is incomplete.}
\label{benchmark}
\end{figure*}

	The result of performance can be seen from the Figure \ref{cpu benchmark} and Figure \ref{gpu benchmark}.
	As shown in Figure \ref{cpu benchmark}, CPU-only \texttt{SPAIC} and \texttt{SpikingJelly} perform best in small
scale network with $n \le 10^5$.
CPU-only \texttt{SpikingJelly} shows the best performance on larger scale networks with $n \ge 10^6$.

	Figure \ref{gpu benchmark} depicts that \texttt{SPAIC} performs best on $n \le 10^6$.
	The \texttt{BrainPy} platform performs well on $n=10^4$, but becomes relatively slow for $n \ge 10^5$ and simulation
runs are stopped for $n \ge 10^7$, due to out of memory.
	Since \texttt{Brian2} simulates spike transmission with event-driven method, the performance of \texttt{Brian2} with
different spike input rate will be different.
Then, we set a moderate spike input rate in benchmark test.
	Figure \ref{gpu benchmark} shows that \texttt{SPAIC} and \texttt{SpikingJelly} perform similarly at large scale.
	
	Although our benchmark are not indicative of the performance in all circumstances, it still shows the performance
of \texttt{SPAIC} is superior to existing platforms.
	Meanwhile, the rich functionality shown in the Table \ref{FrameworkComparison} is also one of the highlights of
\texttt{SPAIC}.
In addition, the flexibility is a significant dimension for evaluating platforms.
We define the flexibility according to the ability of the framework to develop learning algorithms.
As show in the Table \ref{FrameworkComparison}, the deep learning frameworks support gradient back propagation
learning algorithm and the computational neuroscience frameworks support spike timing dependent learning algorithm
(STDP).
Currently, only \texttt{SPAIC} and \texttt{BindsNET} support two types of learning algorithms.

We are now able to characterize the typical platforms in terms of performance, flexibility, and functionality, demonstrating that \texttt{SPAIC} is superior to its counterparts.
\begin{figure}[htbp]
    \centering
        \includegraphics[scale=.15]{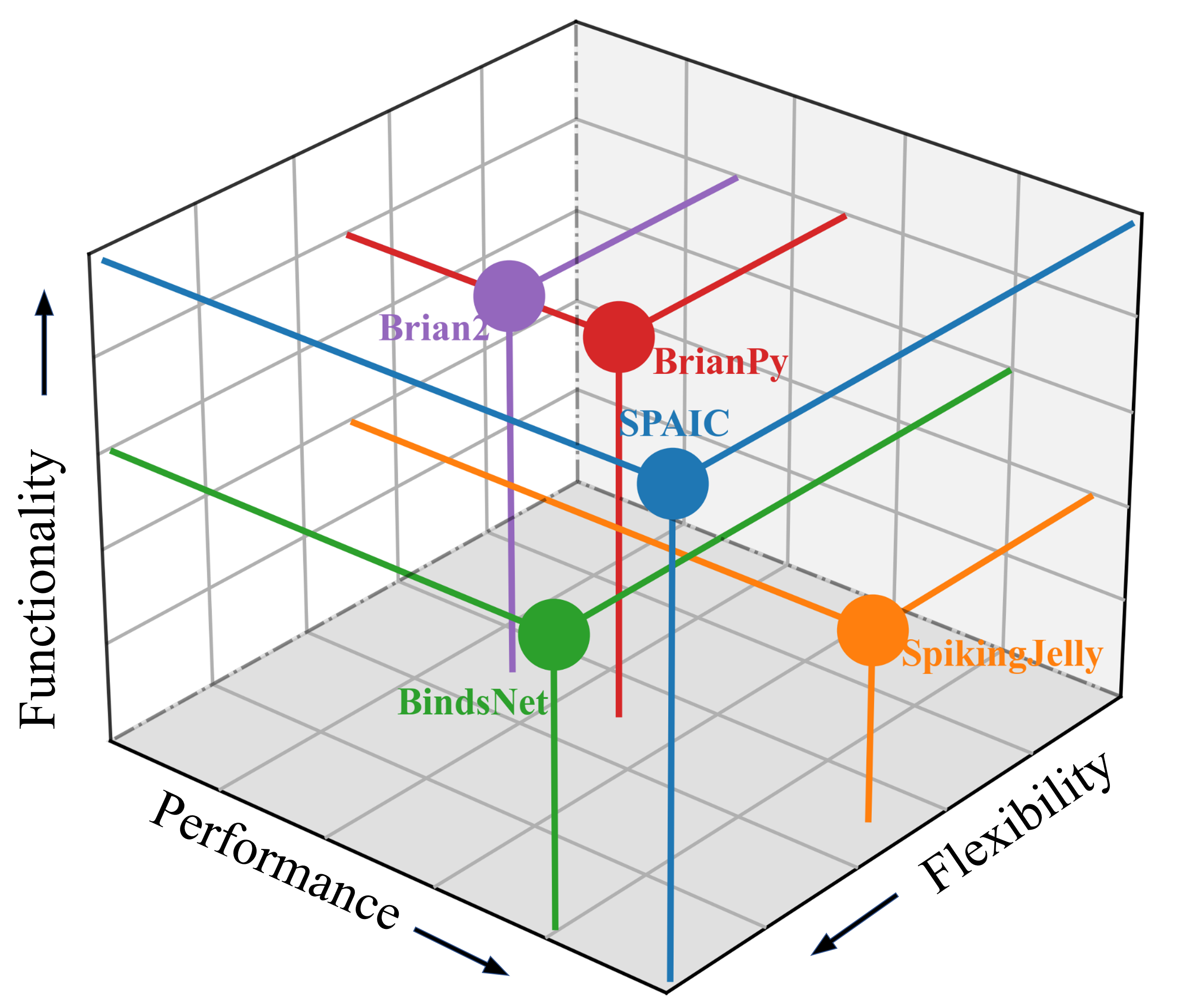}
    \caption{Comparison of performance, flexibility and functionality of frameworks mentioned above.}
    \label{benchmark_3d}
\end{figure}

\section{Further development}

The \texttt{SPAIC} framework is under continuing development, and there is still much room for improvement.
Below we listed several ongoing developments and future directions of the framework:
\begin{enumerate}
    \item More algorithm and model support:
    We are continuously extending our support of SNN learning algorithms, encoding/decoding methods and
    neuronal/synaptic models.
    We plan to integrate more learning types, such as genetic algorithm and network architecture search.
    In the future, we will also add new models and algorithms developed in our research group based on the \texttt{SPAIC}.
    Even though users would most likely use their own customized models and algorithms in their application, a rich
    support of current models provides a quick start for working with the framework.
    \item Performance optimization:
    Optimizing the speed and memory usage of training and simulation will improve the efficiency of experiments with
    SNNs.
    The frontend-backend structure allows indepth optimization on performances, but also requires a lot of time to
    refine detailed computation.
    There are several ongoing attempts to improve the performance: 1) In torch backend, we have used Torch.FX to transform
    dynamic computation graph into static IR, and we can further improve computational efficiency by fuse and transform
    operations;
    2) Add \texttt{JAX} to the backend engine, and use its \texttt{JIT} techniques to improve performance;
    3) Customize the representation and computation of sparse matrix to improve the efficiency in large sparse networks.
    \item ODE solver support:
    Currently, the SPAIC uses discrete dynamical models to iteratively compute the models in the form of euler
     or exponential euler method, which is easy for Back-Propagate Through Time (BPTT) learning algorithms.
    However, to support simulation with various precision and support a more explicit representation of dynamical model,
    we plan to support model representation with ODE form, and adding ODE solvers such as Runge-Kutta methods.
    To support gradient-based learning in more complex ODE solvers, adjoint methods should be imported to the framework.
    \item Extending tools and GUI:
    In this primary version of \texttt{SPAIC}, we only provide core functions of SNN simulation and training, but we plan to
    enrich our framework with tools that may be useful in model building and analysis.
    For example, we plan to support visualization tools like \texttt{Tensorboard} into the framework that can help
    analyze the model in aspects such as network structure, dynamics, parameter distribution and neuron selectivity.
\end{enumerate}

\section{Discussion}
The intention of developing the \texttt{SPAIC} framework is to assist new brain-inspired modeling study and various
application research in the neuromorphic community, including spiking neural networks algorithms, neural system modeling,
and robotic systems, etc.

Neuromorphic computing is still an emerging multidisciplinary research field, where multiple theories and
methodologies are competing and integrating with each other, and the boundary of this field is still not very clear.
Hence, it imposes a remarkable challenge to decide what features should be emphasized for such a computing framework,
supporting both training and simulation.
Many researchers viewing SNN as a special RNN, such approach provides a shortcut for development and implementation of
SNN algorithms.
However, it could also pose limitations or even be misleading for brain-inspired computing researches, as there are
fundamental differences between SNNs and RNNs.
For example, RNN populations are designed to compute output at every time step where the information processing occurs at
population level (tensor) while spiking neurons only emit spikes very sparsely and computation mainly
happens within single neuron dynamics \cite{gerstner2002spiking}.
Several recent works have developed algorithms with deep SNNs such as the spiking ResNet, which have
achieved comparable performance with counterpart deep learning algorithms in image classification tasks
\cite{fang2021deep}.
However, to achieve such performance, the SNN have to be simulated with simple neuron model, very short time steps and
very high firing rate, where the spiking neuron can hardly contain memory information or process temporal information
by the neuronal dynamics.
Such approach limited the potential of SNNs both in power efficiency and computation capability.
To facilitate truly brain-inspired computation researches, we designed the \texttt{SPAIC} network construction interface with
neuroscience style such that researchers can easily introduce neuroscience results into brain-inspired models.
On the other hand, deep learning methodologies provide a variety of delicate training algorithms for optimizing network functions,
hence we incorporated the training procedure into our framework, and use a \textit{Learner} object to integrate theories in
the both field into one united procedure.
This hybrid coding style is what we see most suitable for current brain-inspired modeling studies.

To confront foreseeable development of brain-inspired modeling methodologies, we have to seek the balance between the
guidance of neuroscience principles and the deploying of deep learning technologies, and provide considerable
flexibility in the framework.
In \texttt{SPAIC}, we provide a guideline to build a network model by combining assembly, connection and learner
objects, with each object specialized in one aspect of neural computation.
On the other hand, each of those network components provides interfaces to directly customize their backend computation,
and hence users can easily build their model.
Moreover, even through we have provided functions to model biological features such as synaptic dynamics,
sparse connections and conduction delay, users can also easily implement their realizations to fit their usage.
In summary, \texttt{SPAIC} is a highly customizable framework, which can help researchers easily and efficiently
build and test brain-inspired models and develop various artificial intelligence applications.

\section*{Acknowledgments}
The authors would like to acknowledge the National Key Research and Development Program of China under grant 2020AAA0105900,
the Key Research Project of Zhejiang Lab under Grant 2021KC0AC01, and the National Natural Science Foundation of China under Grant
62106234 for financial support.

\clearpage
\printbibliography

\section{Supplementary Information}
\renewcommand{\thefigure}{S\arabic{figure}}
\setcounter{figure}{0}

\begin{figure}[H]
    \centering
    \fbox{\includegraphics[scale=.70]{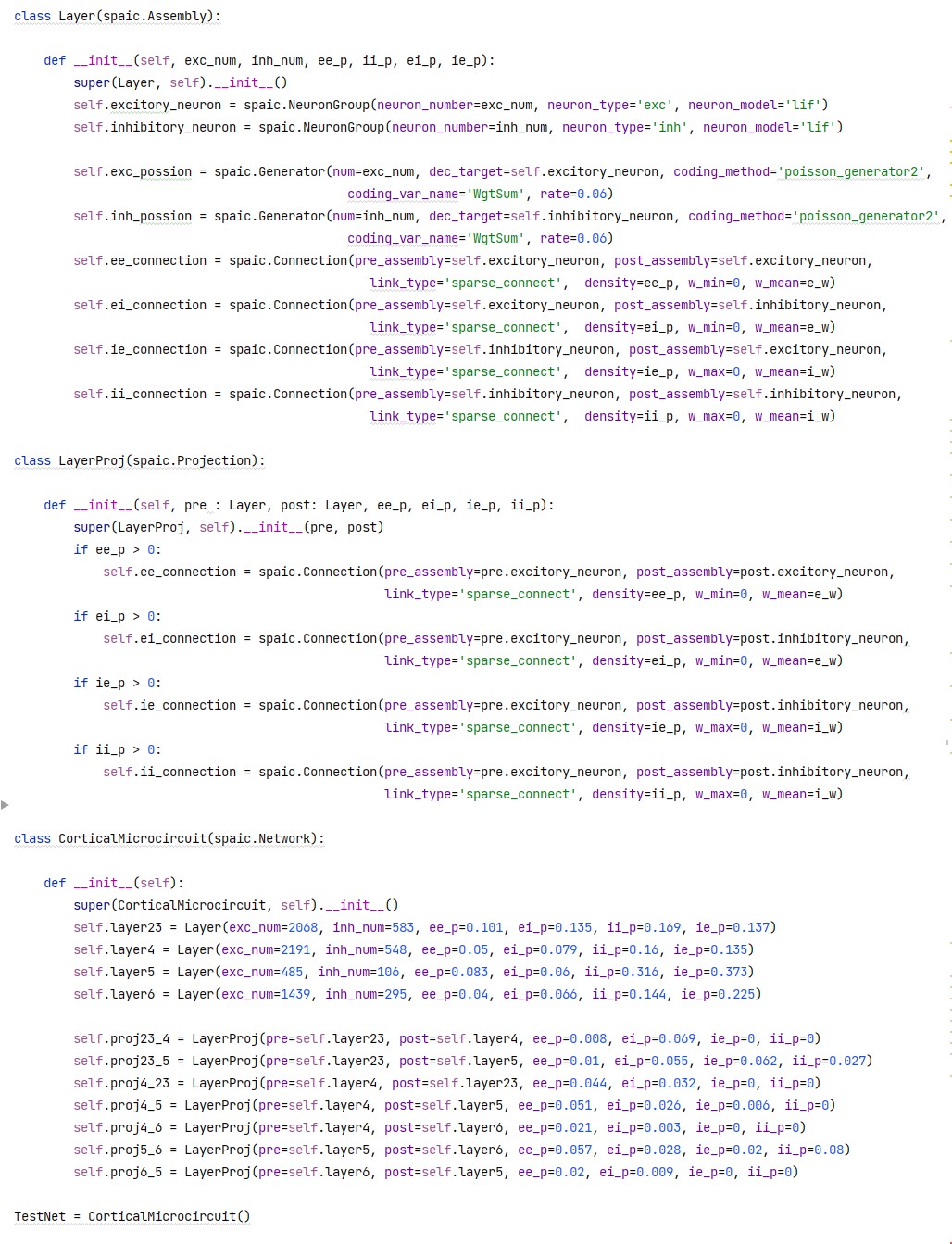}}
    \caption{Example code of cortical network}
    \label{CorticalCode}
\end{figure}

\renewcommand{\thefigure}{S\arabic{figure}}
\begin{figure*}[tbp]
    \centering
    \fbox{\includegraphics[width=.95\linewidth]{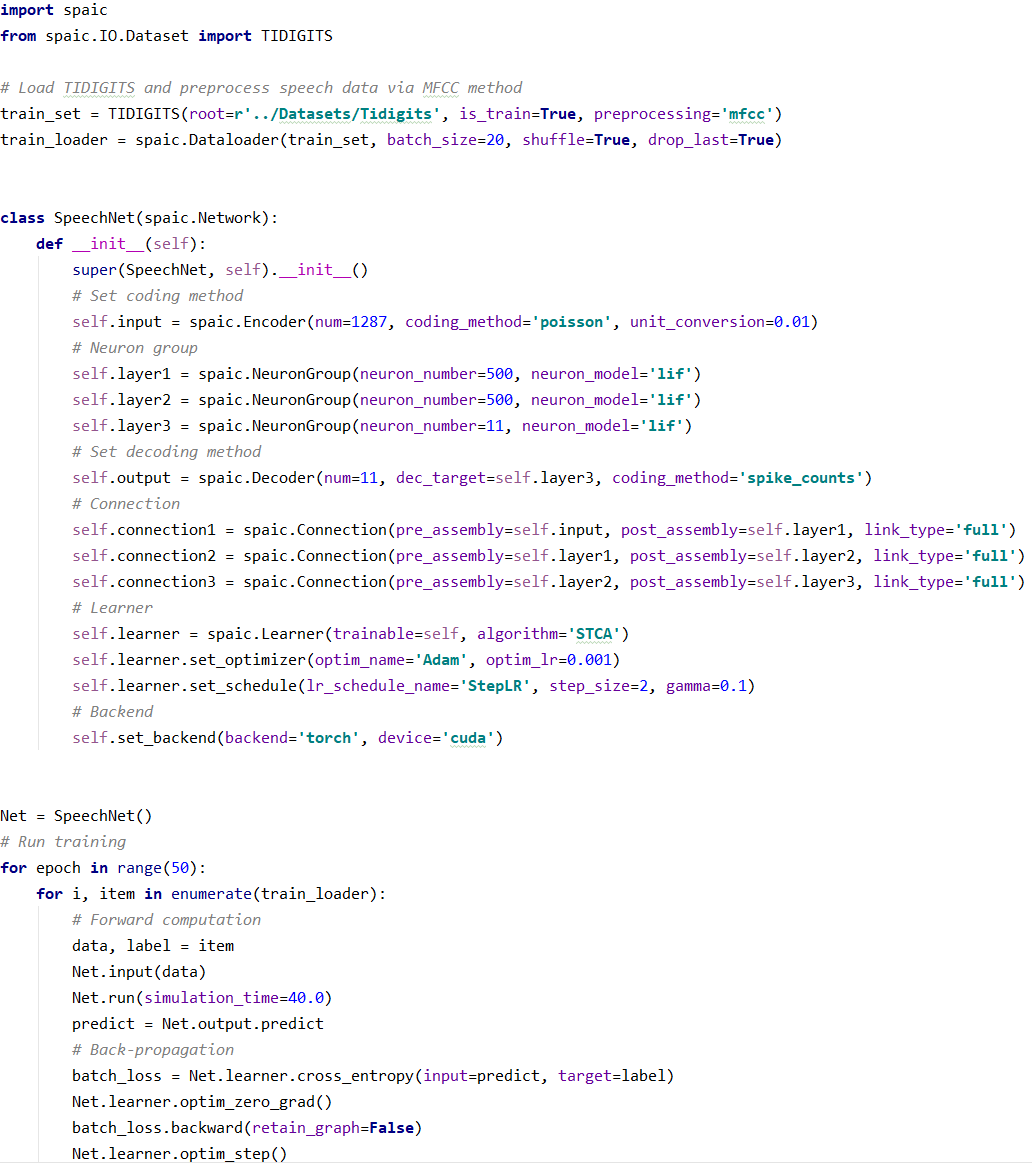}}
    \caption{Example code of speech recognition network.}
    \label{speech_code}
\end{figure*}

\renewcommand{\thefigure}{S\arabic{figure}}

\begin{figure*}[tbp]
    \centering
    \fbox{\includegraphics[width=.95\linewidth]{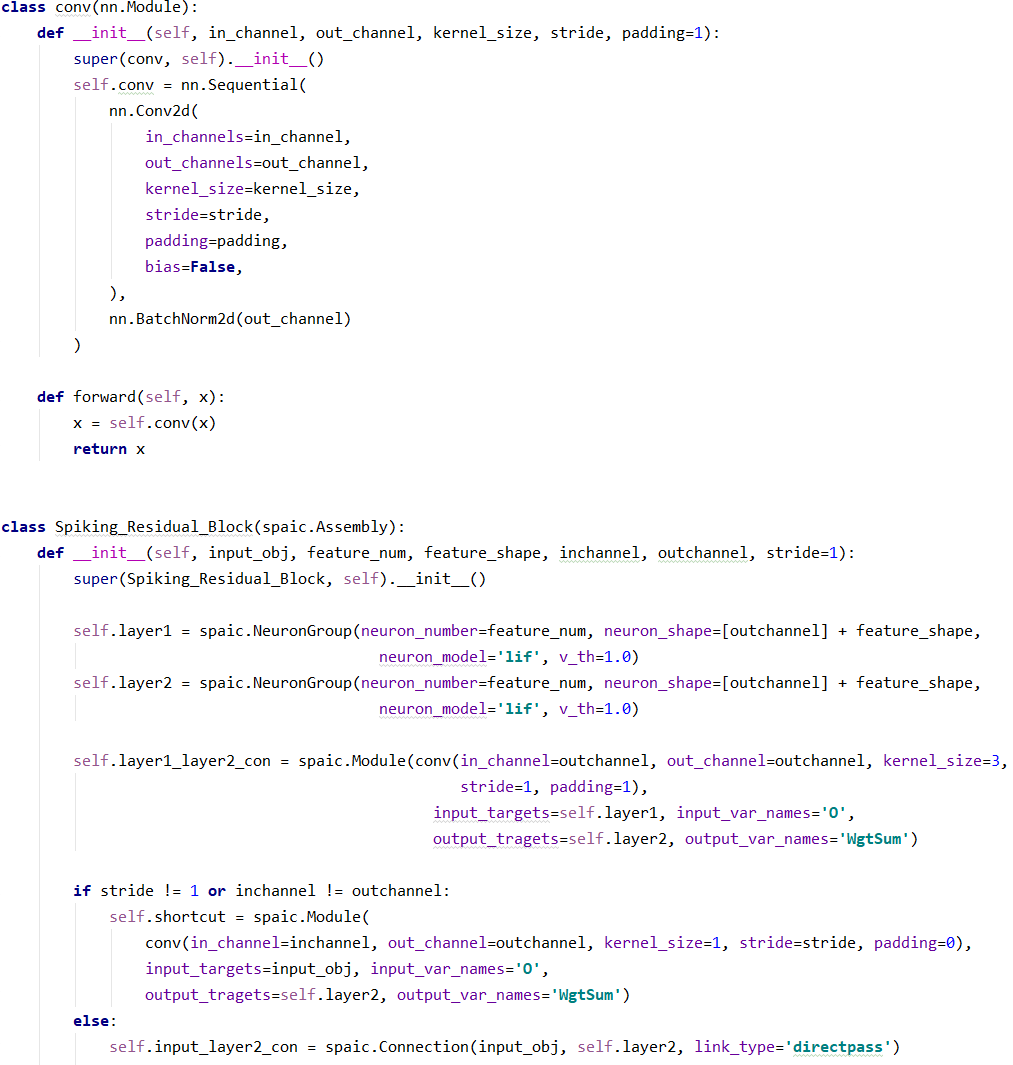}}
    \caption{Example code of residual block.}
    \label{image_code}
\end{figure*}


\end{document}